\RequirePackage[T1]{fontenc}
\documentclass[journal]{IEEEtran}
\usepackage{microtype}
\usepackage{xcolor} % Blue marks identify method revisions.
\usepackage{amsmath,amsfonts,amssymb}
\usepackage{algorithmic}
\usepackage{algorithm}
\usepackage{array}
\usepackage[caption=false,font=normalsize,labelfont=sf,textfont=sf]{subfig}
\usepackage{textcomp}
\usepackage{stfloats}
\usepackage{url}
\usepackage{verbatim}
\usepackage{graphicx}
\usepackage{cite}
\usepackage{booktabs}
\usepackage{multirow}
\usepackage{placeins}
\usepackage{orcidlink}
\begin{document}
\bstctlcite{uniml:bstcontrol}
\raggedbottom

\title{A Unified Framework for Image Manipulation Localization Under Multi-Type Weak Supervision}

\author{Songlin Li\textsuperscript{\large\orcidlink{0009-0008-6793-5624}}, Hui Ma, Zhiqing Guo\textsuperscript{\large\orcidlink{0000-0001-6412-334X}},~\IEEEmembership{Member,~IEEE}, Dan Ma,\textsuperscript{\large\orcidlink{0009-0002-2262-9603}}~\IEEEmembership{Member,~IEEE}, Gaobo Yang\textsuperscript{\large\orcidlink{0000-0003-2734-659X}}, \\ and Xin Liao, ~\IEEEmembership{Senior Member,~IEEE}
        % <-this % stops a space
\thanks{This work was supported in part by the Central Government Guides Local Science and Technology Development Fund Projects under Grant ZYYD2026ZY21, in part by the National Natural Science Foundation of China under Grant 62302427, and Grant 62462060. (Songlin Li and Hui Ma contributed equally to this work.) (Corresponding author: Zhiqing Guo.) 

}% <-this % stops a space

\thanks{Songlin Li, Hui Ma, Dan Ma, and Zhiqing Guo are with the School of Computer Science and Technology, Xinjiang University, Urumqi 830046, China. (e-mail: guozhiqing@xju.edu.cn).

 Gaobo Yang and Xin Liao are with the College of Computer Science and Electronic Engineering, Hunan University, Changsha 410082, China.

}
}

\markboth{Journal of \LaTeX\ Class Files,~Vol.~14, No.~8, August~2021}%
{Shell \MakeLowercase{\textit{et al.}}: A Sample Article Using IEEEtran.cls for IEEE Journals}


\maketitle

\begin{abstract}
 Existing weakly supervised image manipulation localization (IML) methods typically rely on annotation-specific training strategies, making it difficult to jointly use different annotation types within a single model. This limits the scale of usable training data, thereby creating a bottleneck for further progress in weakly supervised IML. To address this challenge, we propose UniIML, a teacher–student framework that maps diverse weak annotations into a unified representation. This representation  guides the Segment Anything Model (SAM) to generate multiple candidate masks. However, SAM is designed for general-purpose segmentation and has limited sensitivity to manipulation traces. Its predicted quality scores alone may fail to reliably identify the candidate that best delineates the manipulated region. We therefore propose Ordinal Safe Consensus (OSC), which first filters out implausible candidates using trusted weak-annotation constraints. OSC then aggregates candidate rankings based on annotation consistency, SAM-predicted quality scores, and cross-view stability, repeating the selection with each criterion omitted in turn. The most frequently selected candidate is adopted as the final mask, and its relative selection frequency is used to quantify selection stability. Then, we construct soft pseudo-labels from the selected candidate’s predicted probabilities and estimate pixel-level supervision weights using selection stability, pixel-level agreement among candidates, and predictive certainty. The student model receives direct supervision in regions defined by the weak annotations and learns from soft pseudo-labels in the remaining regions via reliability-weighted distillation, enabling unified training across weak annotation types. Experiments across nine datasets with four types of weak annotations show that UniIML outperforms the compared weakly supervised methods under both single-type and mixed-type annotation settings, while remaining competitive with several fully supervised methods.
\end{abstract}

\begin{IEEEkeywords}
Image manipulation localization, weak supervision, unified framework, pseudo-labels.
\end{IEEEkeywords}

\section{Introduction}

\IEEEPARstart{T}{he} rapid development of image-editing software and generative models has made visually convincing manipulated content increasingly accessible, raising concerns about misinformation and the credibility of digital media~\cite{verdoliva2020media,mirsky2021creation,kadha2025unravelling}. Image manipulation localization (IML) aims to identify manipulated regions in an image at the pixel level. In recent years, deep learning-based methods have substantially improved IML performance by jointly exploiting complementary forensic cues, including image content~\cite{li2025bridging,niu2026dr}, boundary artifacts~\cite{dong2022mvss,liu2024attentive,yu2026earg}, noise residuals~\cite{guillaro2023trufor,zhang2026noise}, frequency-domain traces~\cite{li2024unionformer,wang2022objectformer}, and 3D geometric cues~\cite{yu20262d}. However, their performance generally relies on pixel-level  masks, whose annotation is expensive because manipulated regions often have irregular shapes, fragmented structures, and ambiguous boundaries~\cite{lin2016scribblesup}. Therefore, the high cost of pixel-level mask annotations has motivated researchers to explore weakly supervised IML.

\begin{figure}[!t]
  \centering
  \includegraphics[width=\linewidth]{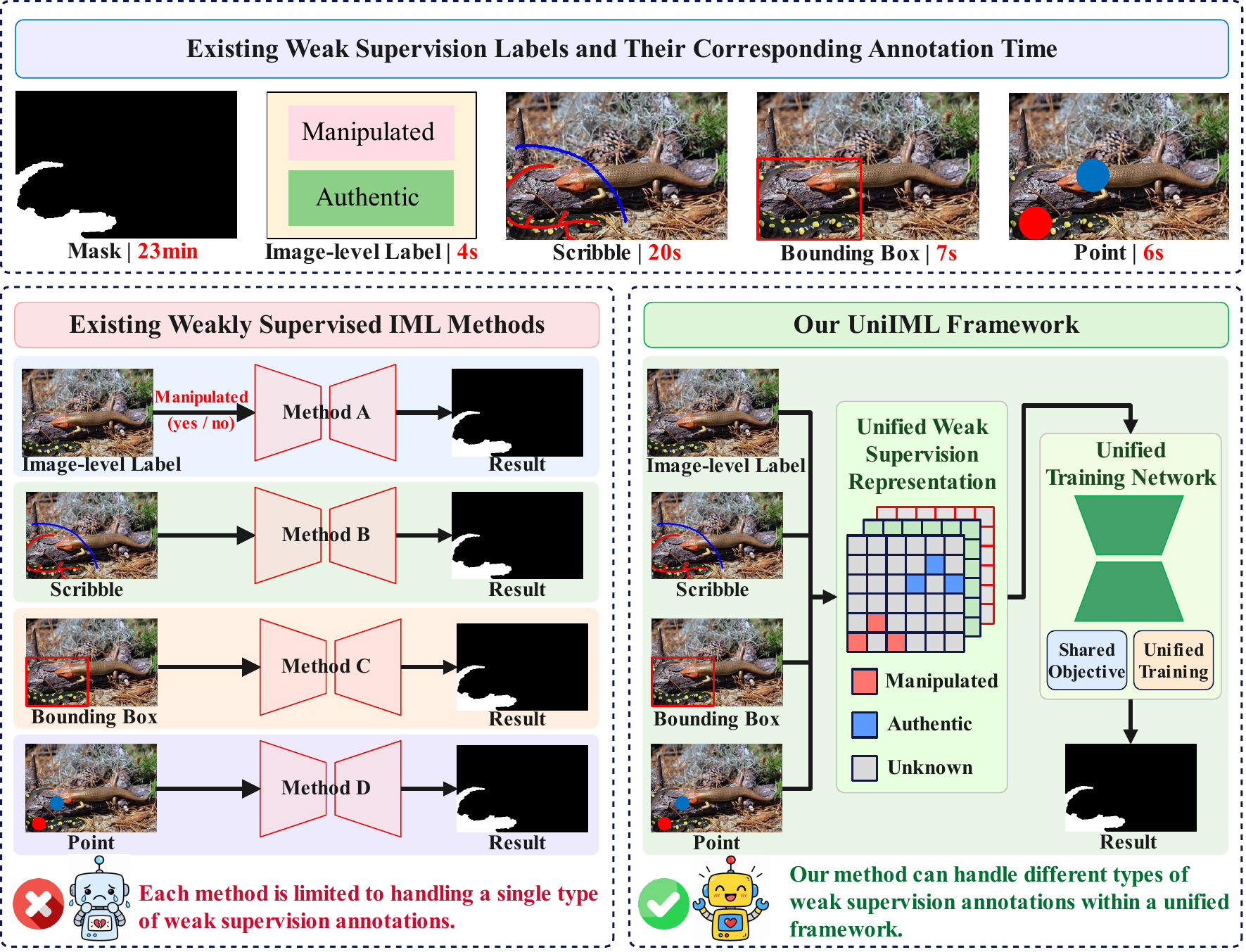}
  \caption{Overview of our unified weak-supervision framework. Unlike existing methods that are typically tailored to a single annotation type, our framework jointly supports image-level labels, bounding boxes, scribbles, and points.}
  \label{fig_1}
\end{figure}

Existing weakly supervised IML methods primarily employ three types of annotations: image-level labels~\cite{ZHOU2024123501,zhai2023towards,bai2025weakly,10889843}, bounding boxes~\cite{guo2026passive}, and scribbles~\cite{li2026beyond}. In this work, we further introduce point annotations as an additional form of weak supervision for IML, as illustrated in Fig.~\ref{fig_1}. Compared with pixel-level masks, these weak annotations can reduce annotation time from minutes to seconds, significantly lowering annotation costs. However, different types of weak annotations provide distinct supervisory information. Specifically, image-level labels indicate only the overall authenticity of an image without providing explicit spatial information about manipulated regions. Point annotations identify the manipulation status of a small number of manually verified locations. Bounding boxes provide the approximate spatial extent of manipulated regions but do not guarantee that all enclosed pixels are manipulated, while scribbles offer local supervision through sparsely labeled foreground or background pixels. Consequently, these annotation types differ substantially in their spatial coverage and the certainty of the supervision they provide. To exploit the characteristics of different weak annotations, researchers have developed various weakly supervised IML methods tailored to specific annotation types. Under image-level supervision, due to the lack of explicit spatial information, existing methods typically employ class activation maps~\cite{ZHOU2024123501}, consistency learning~\cite{zhai2023towards,bai2025weakly}, or self-optimization strategies~\cite{10889843} to discover potential manipulated regions. In contrast, bounding box-supervised~\cite{guo2026passive} and scribble-supervised~\cite{li2026beyond} methods leverage the available spatial constraints to infer the manipulation status of unlabeled regions, thereby improving localization accuracy. Although these methods have achieved promising performance under their respective supervision settings, their designs are typically tailored to specific annotation types, making it difficult to jointly utilize training data with different forms of annotations. This limitation not only restricts the effective utilization of available annotated data but also reduces the flexibility of weakly supervised IML methods in practical applications.  Therefore, a unified learning framework capable of accommodating multiple types of weak annotations is needed. This raises a key question: How can different types of weak annotations be integrated into a unified learning framework while preserving the information provided by each annotation type and avoiding supervision conflicts during joint training?

To address the above problem, our core idea is to map different types of weak annotations into a unified supervisory representation and leverage the Segment Anything Model (SAM)~\cite{kirillov2023segment} to generate pseudo-masks for distilling a student model. However, realizing this idea presents three major challenges. First, different types of weak annotations vary in spatial constraints and supervisory certainty, and ignoring these differences may introduce incorrect or conflicting pixel-level supervision. Thus, the unified representation must preserve the valid constraints of each annotation type. Second, SAM may generate multiple plausible candidate masks under weak annotation constraints, making it crucial to select reliable pixel-level supervisory signals to ensure pseudo-label quality. Third, the reliability differences between human-provided weak annotations and teacher-generated pseudo-labels must be properly addressed. This requires prioritizing trusted human annotations while adaptively adjusting the contributions of different supervisory signals to training, thereby mitigating the adverse effects of ambiguous regions and candidate selection errors on student model learning.

To sum up, we propose UniIML, a unified teacher–student framework for weakly supervised image manipulation localization (IML) that supports four types of weak annotations: image-level labels, points, bounding boxes, and scribbles. UniIML supports both single-type and mixed-type weak annotation settings. The framework encodes these annotations into a unified representation that preserves the supervision provided by each type and enables a shared pseudo-label generation pipeline to supervise a single student model. In the teacher stage, the unified representation guides SAM to generate multiple candidate masks. However, SAM is designed for general-purpose segmentation and has limited sensitivity to manipulation traces. Its predicted quality scores alone may fail to reliably identify the candidate that best matches the manipulated region. To address this limitation, we propose Ordinal Safe Consensus (OSC), which first applies safety filtering to retain candidate masks that best satisfy the trusted foreground, background, and spatial constraints encoded in the unified representation. It then performs ordinal consensus voting by aggregating candidate rankings based on annotation consistency, SAM-predicted quality scores, and cross-view stability. By repeating the selection with each of these three criteria omitted in turn, OSC adopts the most frequently selected candidate as the final mask and uses its relative selection frequency to quantify selection stability. We then generate soft pseudo-labels from the selected candidate and estimate pixel-level reliability weights based on selection stability, agreement among candidates, and predictive certainty. In the student stage, the model receives direct supervision in regions defined by the weak annotations and learns from soft pseudo-labels in the remaining regions through reliability-weighted distillation. By combining annotation-derived supervision with teacher-generated pseudo-labels, UniIML preserves the supervision provided by weak annotations while extending it into dense training signals, enabling joint training across annotation types. We evaluate UniIML on nine IML datasets under four weak-annotation settings and further investigate joint training with mixed annotation types. Experimental results demonstrate its effectiveness across annotation settings and its cross-dataset generalization ability, outperforming the compared weakly supervised baselines while remaining competitive with several fully supervised methods.

This work extends our AAAI 2026 paper~\cite{li2026beyond} in three respects: data, task formulation, and methodology. At the data level, we extend the original scribble-only supervision setting to multiple forms of weak annotation, including image-level labels, point annotations, bounding boxes, and scribbles, thereby covering a broader range of supervision granularities and annotation costs. In particular, to the best of our knowledge, this is the first work to introduce point-level supervision into IML and to systematically investigate weakly supervised manipulation localization based on point annotations. Compared with bounding boxes and scribbles, point annotations require human verification at only a small number of pixel locations, providing a more annotation-efficient form of spatial supervision and establishing a new low-cost learning setting for IML. At the task level, we generalize scribble-supervised IML into a unified weakly supervised IML task that accommodates multiple annotation types and allows images with different forms of supervision to be jointly used for model training within a single framework. At the methodological level, we extend the original scribble-driven approach into a unified teacher--student framework. Different types of weak annotations are represented through a shared constraint formulation, while candidate-mask generation and evaluation, OSC, and confidence-aware optimization are further introduced to improve the reliability of pixel-level supervision and enhance the model's ability to learn consistently from multiple forms of weak supervision. 

The main contributions of this work are summarized as follows:

\begin{itemize}
\item We propose UniIML, which unifies IML under image-level, point, bounding-box, and scribble supervision within a single teacher-student framework. Through a unified supervision representation, UniIML preserves the information provided by each annotation type while enabling a single IML model to learn under both single-type and mixed-annotation settings.

\item We propose Ordinal Safe Consensus (OSC), a candidate-mask selection mechanism that combines safety filtering with consensus voting based on relative rankings. By aggregating within-image candidate rankings rather than directly fusing evidence scores on different scales, OSC avoids tunable evidence-fusion weights and additional learned score calibration. It further assesses selection stability through the consistency of candidate selection when each evidence group is omitted in turn.

\item We develop a reliability-aware student learning strategy that combines direct supervision in regions defined by weak annotations with reliability-weighted soft pseudo-label distillation in the remaining regions. The reliability weights integrate selection stability, agreement among candidates, and predictive certainty, extending supervision to unlabeled regions while preventing uncertain teacher predictions from overriding known labels.

\item We conduct extensive experiments across nine IML datasets under both single-type and mixed-annotation settings involving four types of weak annotations. The results demonstrate the effectiveness of UniIML in utilizing different annotation types, outperforming the compared weakly supervised methods while remaining competitive with several fully supervised methods. 
\end{itemize}

\section{Related Work}
Different forms of supervision vary in spatial information, annotation cost, and localization strategies. This section reviews fully supervised and weakly supervised IML methods relevant to the proposed approach. Fully supervised methods rely on pixel-level manipulation masks that provide precise region boundaries but incur high annotation costs. In contrast, weakly supervised methods use incomplete annotations, such as image-level labels, bounding boxes, points, or scribbles, to reduce annotation costs while requiring models to infer missing spatial information. Thus, the two paradigms differ in supervision granularity, reliability, and how manipulated regions are localized.

\subsection{Fully Supervised Image Manipulation Localization}
A central challenge in fully supervised IML is to distinguish manipulation-induced artifacts from natural variations in image content. One line of research therefore focuses on manipulation-sensitive low-level traces and the integration of complementary forensic cues. ManTra-Net~\cite{wu2019mantra} learns manipulation-sensitive features through image-operation classification and formulates localization as local anomaly detection, whereas CAT-Net~\cite{kwon2021cat} exploits compression artifacts by jointly modeling features in the RGB and discrete cosine transform (DCT) domains to identify spliced regions. These approaches exemplify the use of operation-related traces and compression statistics as forensic evidence. Building on the complementarity of different cues, MVSS-Net~\cite{dong2022mvss} jointly exploits noise information and manipulation-boundary artifacts, with multi-scale supervision supporting both localization and detection. TruFor~\cite{guillaro2023trufor} further integrates RGB content with learned noise-sensitive fingerprints and estimates the reliability of pixel-level predictions. Together, these methods demonstrate how combining forensic evidence from different sources can reduce dependence on any single visual cue.
Complementary to forensic cue extraction and fusion, another line of research investigates relationships across spatial locations and scales to recover manipulated regions more completely. SPAN~\cite{hu2020span} models relationships among image patches at multiple scales using a pyramid of local self-attention operations, while PSCC-Net~\cite{liu2022pscc} combines spatio-channel correlations with progressive multi-scale prediction to recover manipulation masks from coarse to fine. Rather than making decisions from isolated local features, these methods interpret forensic evidence within broader spatial relationships. Extending this direction, ObjectFormer~\cite{wang2022objectformer} jointly processes RGB and high-frequency features and uses learnable object prototypes to model object- and patch-level consistency. Mesorch~\cite{zhu2025mesoscopic} combines convolutional and transformer representations within a multi-scale framework, connecting fine-grained forensic details with global semantic information. Through relational modeling and hybrid architectures, these studies explore how local anomalies and broader context can jointly support localization. Beyond expanding contextual modeling, recent studies have also examined how semantic information and prior knowledge should be selectively incorporated into forensic representations. SparseViT~\cite{su2025sparsevit} replaces dense global self-attention with sparse interactions to reduce reliance on image semantics and adaptively extract manipulation-related nonsemantic features. Other approaches introduce auxiliary priors to guide forensic learning. PNF-Net~\cite{li2026prototype} organizes manipulation traces into reusable prototype-memory priors, whereas CMB-Net~\cite{li2025bridging} incorporates textual semantics generated by a language model to enrich visual representations through fine-grained vision–language interactions, while weighting potentially ambiguous textual information. These different strategies highlight that the objective is not simply to strengthen or suppress semantics, but to exploit information relevant to manipulation discrimination while limiting interference from irrelevant content or unreliable priors.

Overall, fully supervised methods have advanced forensic representation learning through complementary cue fusion, cross-scale relational modeling, and the integration of semantics and prior knowledge. However, they generally rely on pixel-level manipulation masks for localization supervision. In contrast, UniIML focuses on constructing reliable training supervision when complete pixel-level annotations are unavailable and weak annotations vary in semantics and spatial constraints, thereby complementing existing advances in forensic representation learning.

\subsection{Weakly Supervised Image Manipulation Localization}

Weakly supervised IML aims to reduce reliance on complete pixel-level masks by learning to localize manipulated regions from incomplete supervision. Image-level supervision requires only an authenticity label for each image, without explicitly annotating manipulated regions, and therefore incurs a low annotation cost. Its central challenge is to translate global authenticity supervision into spatially meaningful manipulation responses. Existing methods address this challenge by extracting localization cues through class activation maps~\cite{ZHOU2024123501}, cross-source and inter-patch relationship modeling~\cite{zhai2023towards,bai2025weakly}, and self-optimization strategies~\cite{10889843}. Activation-based approaches identify local evidence supporting image-level classification, whereas relational modeling and iterative refinement exploit dependencies among predictions or regions to compensate for missing spatial supervision. For example, WSCL~\cite{zhai2023towards} combines cross-source pseudo-label refinement with inter-patch consistency learning to extract localization information using only binary authenticity labels.
Under the same supervision setting, several studies further leverage segmentation and semantic priors from pretrained models to improve spatial inference from image-level labels. CAM-guided SAM refinement~\cite{wang2025context} uses class activation responses to guide a pretrained segmentation model, while ViLaCo~\cite{sheng2026weakly} introduces external semantic guidance through vision--language collaborative reasoning. Nevertheless, whether localization relies on internally modeled forensic relationships or external pretrained priors, image-level labels do not directly specify which pixels are manipulated or how far the manipulated regions extend. The resulting spatial supervision therefore remains model-inferred, and image-level labels alone cannot ensure its agreement with the true manipulated regions. The key limitation thus lies in the absence of explicit spatial annotations that directly constrain localization, rather than a lack of image-level authenticity information.
Spatially constrained weak annotations offer an alternative way to address this limitation, providing more explicit spatial information at the expense of additional annotation effort. Existing IML studies have primarily explored bounding boxes and scribbles. BoxPromptIML~\cite{guo2026passive} uses bounding boxes to indicate the approximate extent of manipulated regions and to prompt a frozen SAM teacher, subsequently transferring the resulting localization knowledge to a student network. In contrast, our previous scribble-supervised framework~\cite{li2026beyond} learns directly from sparsely annotated manipulated and authentic pixels, while exploiting the remaining unlabeled regions through structural consistency and confidence-aware regularization. These two annotation types reduce spatial ambiguity in different ways: bounding boxes constrain the spatial extent of manipulated regions without implying that every enclosed pixel is manipulated, whereas scribbles provide local pixel labels without delineating the complete regions. 

These studies reflect different trade-offs between annotation cost and the strength of spatial supervision. However, existing methods are typically tailored to a particular annotation type, focusing on how to recover missing pixel-level supervision within that specific setting. When training data contain different forms of weak annotation, sharing a localization network alone is insufficient to unify the supervision process, because image-level labels, points, bounding boxes, and scribbles differ in the information they establish, their spatial coverage, and the interpretation of unlabeled regions. In contrast to annotation-specific approaches, UniIML focuses on constructing reliable pixel-level training supervision within a unified formulation while preserving these distinctions. This enables different annotation types to be used individually or jointly within a single teacher--student framework.

% \subsection{Promptable Segmentation and Reliability-Aware Distillation}
\begin{figure*}[t]
  \centering
   \includegraphics[width=\linewidth]{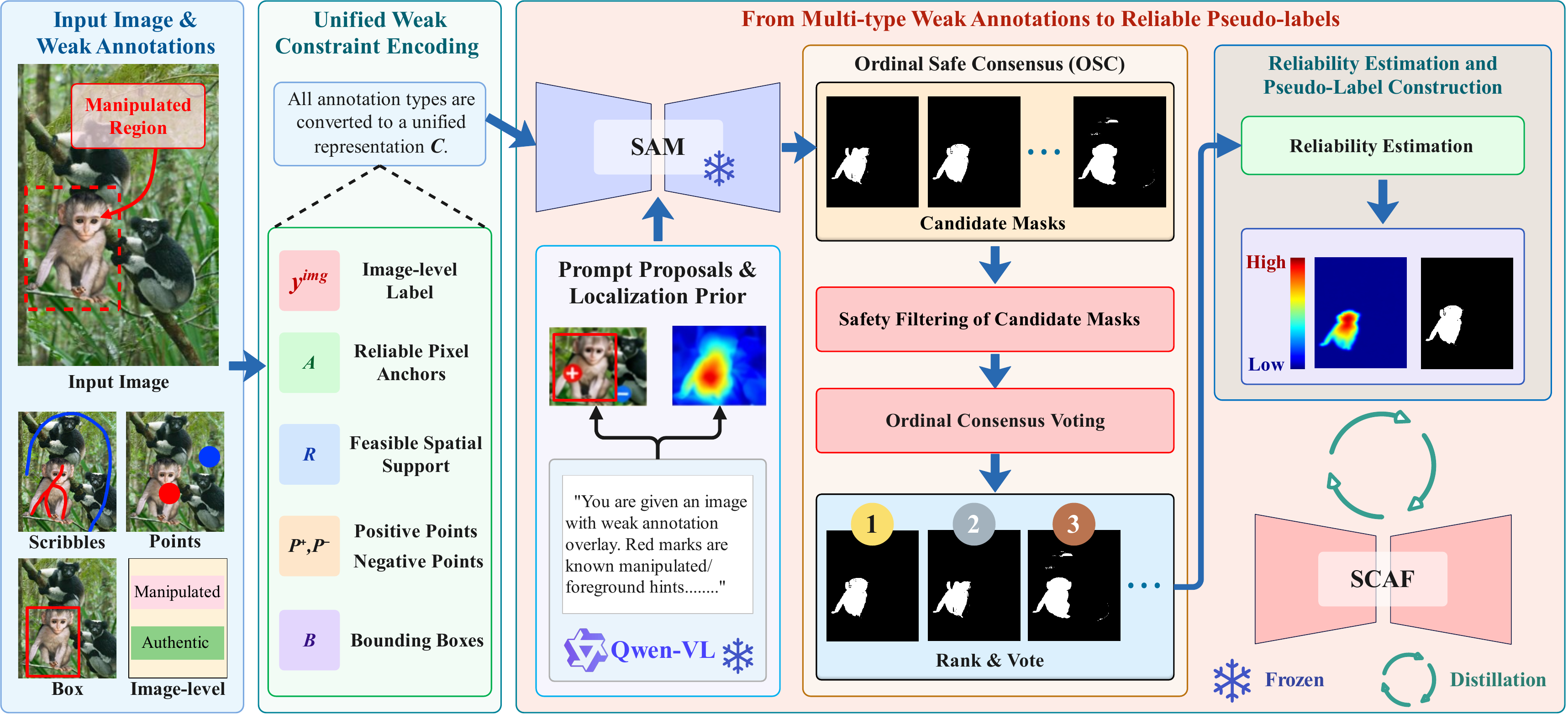}

\caption{Overview of UniIML. Unified weak annotations, vision-language prompts, and localization priors guide a frozen SAM to generate candidate masks. OSC selects a candidate mask and assesses its selection stability. The student is trained using reliability-weighted soft pseudo-labels, direct annotation supervision, and a weighted soft Dice loss.}
   \label{fig:2}
\end{figure*}

% Method-only fragment; requires amsmath, amsfonts, amssymb and xcolor.
% Retain the original dataset table and the overview figure when integrating.
% The full-manuscript copy already preserves these floats.
\section{Methodology}
\label{sec:method}

\subsection{Overview}
UniIML addresses image manipulation localization (IML) under image-level label, point, bounding-box, and scribble supervision. The training set may use a single annotation type or contain different annotation types across images. As illustrated in Fig.~\ref{fig:2}, UniIML adopts a teacher–student framework. The teacher encodes weak annotations into a unified representation through a shared interface and integrates it with Qwen-VL proposals and localization priors to prompt a frozen SAM, which generates multiple candidate masks. Ordinal Safe Consensus (OSC) selects the final candidate through safety filtering and ordinal consensus voting while quantifying selection stability. The selected candidate’s logits yield dense soft pseudo-labels, with pixel-level supervision weights determined by selection stability, agreement among retained candidates, and predictive certainty. We adopt SCAF, previously introduced in our conference paper~\cite{li2026beyond}, as the student model. Student training combines reliability-weighted soft pseudo-label distillation, direct supervision in annotation-defined regions, and a weighted soft Dice loss.

\subsection{Unified Weak Constraint Encoding and Candidate Mask Generation}
\label{subsec:weak_constraints_prompts}

Different weak annotation types convey different pixel-level and spatial
information. Positive image-level labels do not identify manipulated
locations, bounding boxes do not establish every enclosed pixel as
foreground, and points and scribbles label only local pixels. To support
a shared supervision and candidate-generation pipeline, we encode the
information available from each annotation type into a unified
representation.

Given an image $x$ and a weak annotation $a$ of type $t$, the encoder
constructs a unified representation
$\mathcal{C}=\operatorname{Enc}(a,t)$ over the image pixel domain
$\Omega$. This representation contains a partial supervision map $A$,
a spatial support region $R$, labeled point prompts $P$, and box
prompts $B$. Here, $A_i\in\{0,1,\varnothing\}$ denotes background,
foreground, or unknown, respectively, while $R\subseteq\Omega$ guides
prompt placement and the screening of point and box proposals.
These quantities serve distinct purposes: $A$ records pixel-level
supervision targets, whereas $R$ provides a reference region for
prompt search without assigning foreground labels to all enclosed
pixels. SAM~\cite{kirillov2023segment} generates segmentations from
spatial prompts such as points, boxes, and coarse masks. The following
encoding rules convert different annotations into supervision targets
and suitable prompt inputs.

Authentic images receive all-background supervision without candidate
generation. For manipulated images with only positive image-level
labels, all pixels remain unknown in $A$, $R=\Omega$, and no spatial
prompts are directly available from the annotation.
For point annotations, labeled coordinates directly provide point
prompts. To extend single-pixel supervision under a local
label-consistency assumption, we assign foreground to radius-$r$
positive-point disks and background to portions of negative-point
disks not covered by positive disks; other pixels remain unknown.
An expanded box enclosing the positive points serves as a box prompt
and an initial support region. Excluding negative-point disks yields
$R$; without positive points, the initial region is the entire image. For scribble annotations, foreground and background stroke labels
are normalized and aligned to the image resolution to construct $A$.
To obtain SAM-compatible inputs, we sample positive and negative
point prompts from the corresponding strokes and use an expanded
foreground-stroke bounding box as a box prompt and initial support
region. Excluding background strokes yields $R$; without foreground
strokes, the initial region is the entire image. For bounding-box annotations, the boxes serve as box prompts, and
their union defines $R$. Under the complete-coverage assumption,
pixels outside this union are labeled background, while those inside
remain unknown. Box centers provide additional positive point prompts
without receiving foreground labels in $A$.

Annotation-derived prompts can use only the spatial information
already supplied by the annotations, and positive image-level labels
provide no spatial coordinates. To propose additional location
hypotheses, we use a frozen Qwen-VL model~\cite{bai2025qwen3} to
generate point and box proposals from the image and its annotation
context. The model receives the image overlaid with annotation cues,
together with the annotation type, image dimensions, and coordinates.
Proposed coordinates are rounded and clipped to the image bounds.
We then check for conflicts with the encoded labels, assess box size
and overlap with $R$, and remove duplicate proposals. The resulting
points and boxes provide segmentation hypotheses rather than
pixel-level supervision. To complement point and box prompts with per-location guidance, we
further construct a dense response map $H$. When Qwen hidden states
are available, we extract image-token feature vectors from different
network layers, use their norms as scalar responses, and organize
them into coarse spatial response maps. This conversion requires no
additional trainable localization head. The resulting values are used
only to construct prompts and are not interpreted as manipulation
probabilities. To incorporate both encoded information and local image variations,
we select response maps using the available foreground--background
response differences, response differences inside and outside $R$,
and correlations with an image-based reference response. This
reference response is constructed from image residuals, edges, and
local-contrast cues. The selected maps are averaged and combined with
the reference response, after which fixed rules incorporate encoded
labels, boundary cues, and spatial-support information to obtain $H$.
When Qwen hidden states are unavailable, the reference response
provides the initial response for the same subsequent processing. We then sample positive and negative points from high- and
low-response regions of $H$, respectively, extract boxes from
connected components of a thresholded response map, and construct
a dense mask prompt from $H$. Automatically generated prompts do
not modify the supervision targets already recorded in $A$.

Finally, annotation-derived prompts, Qwen-VL proposals, and prompts
constructed from $H$ form the prompt collection $\mathcal{P}$,
which is supplied to a frozen SAM. Because a single prompt may admit
multiple segmentation interpretations, we use multimask mode and
retain the nonempty binary outputs to form
$\mathcal{M}=\{m_k\}_{k=1}^{K}$, where
$m_k\in\{0,1\}^{|\Omega|}$. Each candidate retains its prompt
identifier, SAM-predicted quality score, and available low-resolution
logits.

\subsection{Ordinal Safe Consensus}
\label{subsec:ordinal_safe_consensus}

Unified prompts enable SAM to generate multiple candidate masks. However, SAM is designed for general-purpose segmentation, and its predicted quality scores do not directly reflect the agreement between candidate masks and manipulated regions. We therefore evaluate candidates using three complementary criteria. Annotation compatibility measures consistency with the encoded labels and spatial support. SAM-predicted quality reflects the model's estimate of mask quality. Cross-view stability assesses consistency under horizontal flipping and agreement among candidates. Since these criteria capture different aspects of candidate reliability, directly combining their raw scores requires specifying their relative contributions. To avoid this dependency, we propose Ordinal Safe Consensus (OSC), which consists of two operations, Safety Filtering and Ordinal Consensus Voting. Safety Filtering determines candidate eligibility based on the original annotation constraints, ensuring that high scores on other criteria cannot compensate for explicit annotation conflicts. Ordinal Consensus Voting then aggregates the relative rankings of eligible candidates across the three criteria. By considering both full-evidence and leave-one-evidence-out voting, it selects the final candidate and evaluates selection stability.

\textbf{Safety Filtering.}
Filtering uses the original annotation information retained in the
unified representation $\mathcal{C}$. Let $F_1$ and $F_0$ denote
the foreground and background pixel sets established by the
annotations. Point annotations contribute only their original
centers, scribbles contribute their normalized and spatially aligned
labeled pixels, and bounding boxes contribute outside-background
labels under the complete-coverage assumption. The set
$\mathcal{B}_{\mathrm{ann}}$ contains original positive annotation
boxes. Each box requires some foreground within it, without
specifying a foreground center or an area fraction. Expanded point
disks, inferred boxes, and automatically generated prompts do not
add annotation facts. For a manipulated image, we activate a separate
foreground-presence requirement, denoted by $e=1$, only when both
$F_1$ and $\mathcal{B}_{\mathrm{ann}}$ are empty. Otherwise, $e=0$,
since foreground existence is already implied by a labeled foreground
pixel or a positive annotation box.

Filtering operates on a nonempty candidate collection at the native
image resolution. For candidate $m_k$, we define pixel-label,
box-presence, and image-presence violations as
\begin{equation}
\begin{aligned}
\nu_k^{(c)}
&=\frac{\sum_{i\in F_c}|m_{k,i}-c|}{|F_c|},
\qquad c\in\{0,1\},\\
\nu_k^{\mathrm{box}}
&=\frac{1}{|\mathcal{B}_{\mathrm{ann}}|}
\sum_{b\in\mathcal{B}_{\mathrm{ann}}}
\mathbf{1}\!\left[\sum_{i\in b}m_{k,i}=0\right],\\
\nu_k^{\mathrm{img}}
&=\mathbf{1}\!\left[\sum_{i\in\Omega}m_{k,i}=0\right].
\end{aligned}
\label{eq:osc_violation_rates}
\end{equation}
where $c=1$ and $c=0$ denote the miss rate of labeled foreground pixels and the false-positive rate of labeled background pixels, respectively.  The indicator function $\mathbf{1}[\cdot]$
equals one when the enclosed condition holds and zero otherwise.
The box violation rate is the fraction of annotation boxes without
predicted foreground. Pixel- and box-based constraints are enabled
only when their corresponding sets are nonempty, whereas the
image-presence constraint requires $e=1$. Let $\mathcal{J}$ index
the enabled constraint families. The strictly feasible candidate set is
\begin{equation}
\mathcal{K}_0
=\left\{k:\nu_k^{(j)}=0,\ \forall j\in\mathcal{J}\right\}.
\label{eq:osc_strict_feasible}
\end{equation}

If $\mathcal{K}_0$ is nonempty, all its candidates are retained.
Otherwise, no supplied candidate satisfies every constraint, requiring
an explicit relaxation. To avoid summing violations with different
semantics, we rank the $K$ candidates separately within each constraint
family in ascending violation order, assigning average ranks to ties.
Let $r_{k,\uparrow}^{(j)}$ denote the resulting rank. For $K>1$
and a nonconstant violation family, its normalized rank is
$\widetilde{r}_k^{(j)}
=(r_{k,\uparrow}^{(j)}-1)/(K-1)$.
When $K=1$ or all candidates have the same violation rate in a
family, its normalized rank is set to zero because it provides no
relative discrimination. Any positive violation still counts when
testing strict feasibility. The retained set is
\begin{equation}
\mathcal{K}_{\mathrm{safe}}=
\begin{cases}
\mathcal{K}_0,
&\mathcal{K}_0\neq\varnothing,\\[3pt]
\displaystyle\operatorname*{arg\,min}_{1\leq k\leq K}
\max_{j\in\mathcal{J}}\widetilde{r}_k^{(j)},
&\mathcal{K}_0=\varnothing.
\end{cases}
\label{eq:safe_candidates}
\end{equation}

The relaxed branch retains all candidates whose worst relative
violation rank is minimal, without constraint-fusion weights or an
additional retention threshold. It neither guarantees annotation
satisfaction nor minimizes total violation magnitude. If no
constraints are available, all candidates are retained. Under
positive image-level supervision alone, the presence requirement
cannot further distinguish the nonempty candidates generated in
the preceding stage.

\textbf{Ordinal Consensus Voting.}
After eligibility is determined, candidates are compared within the
fixed set $\mathcal{K}_{\mathrm{safe}}$ using three scores:
\begin{equation}
\begin{aligned}
z_k^{(\mathrm{con})}
&=\min\!\left(\eta_k^+,\eta_k^-,\eta_k^R\right),\\
z_k^{(\mathrm{qua})}
&=q_k^{\mathrm{SAM}},\\
z_k^{(\mathrm{sta})}
&=\min\!\left(\alpha_k,\beta_k\right).
\end{aligned}
\label{eq:osc_evidence_axes}
\end{equation}
where $\eta_k^+$ and $\eta_k^-$ are the fractions of pixels correctly
predicted within the foreground- and background-labeled regions of
$A$, respectively, and $\eta_k^R$ is the fraction predicted as
background outside $R$. An empty evaluation region contributes a
value of one, so missing information does not reduce compatibility.
This score uses the encoded supervision and spatial cues in $A$
and $R$, including annotation-derived inferences, and thus serves a
different role from the original annotation constraints used for
eligibility. The score $q_k^{\mathrm{SAM}}$ is SAM's predicted mask
quality. Taking the minimum within the compatibility and stability
scores limits compensation between their constituent signals.

Cross-view stability combines flip consistency $\alpha_k$ and
inter-candidate agreement $\beta_k$. To compute $\alpha_k$, we
horizontally flip the image and its prompt, repeat prediction, and
align the outputs back to the original image. For each original
output, we take its maximum intersection over union (IoU) with the
aligned outputs, then average these values over outputs from the
same prompt. Candidates generated by that prompt share this score.
The agreement $\beta_k$ is the mean IoU between $m_k$ and the other
masks in the original candidate collection, with a value of one
when that collection contains a single candidate. Neither score is
recomputed after safety filtering. To compare the relative merits of the three criteria, we convert
each score into a normalized within-image ordinal score. Let
$\mathcal{E}=\{\mathrm{con},\mathrm{qua},\mathrm{sta}\}$ and
$K_s=|\mathcal{K}_{\mathrm{safe}}|$. For criterion
$d\in\mathcal{E}$, retained candidates are ranked in descending
score order, with rank one assigned to the highest score and average
ranks assigned to ties. We define
\begin{equation}
\rho_k^{(d)}=
\begin{cases}
\dfrac{K_s-r_k^{(d)}}{K_s-1},
&K_s>1,\\[5pt]
1,&K_s=1.
\end{cases}
\label{eq:osc_ordinal_scores}
\end{equation}

To examine whether selection depends on a particular criterion,
we use the complete three-criterion combination and three additional
combinations that each omit one criterion. Let $\mathcal{Q}$ denote
these four combinations. Each combination averages its included
ordinal scores with equal weights and selects the highest-scoring
candidate:
\begin{equation}
b_k^{(q)}
=\frac{1}{|q|}\sum_{d\in q}\rho_k^{(d)},
\qquad
k_q=\operatorname*{arg\,max}_{k\in\mathcal{K}_{\mathrm{safe}}}
b_k^{(q)}.
\label{eq:borda_score}
\end{equation}
This aggregation compares relative ranks without learned score
calibration or tunable fusion weights between the three criteria.
The four selections change only the criteria included in the mean;
the candidate set and individual criterion rankings remain fixed.
Consequently, omitting a criterion cannot readmit a filtered
candidate. Each combination casts one vote. The candidate receiving the most
votes is selected, and its vote share defines selection stability:
\begin{equation}
\begin{aligned}
v_k=\sum_{q\in\mathcal{Q}}\mathbf{1}[k=k_q],\quad
\hat{k}=\operatorname*{arg\,max}_{k\in\mathcal{K}_{\mathrm{safe}}}v_k,\quad
s_{\hat{k}}=\frac{v_{\hat{k}}}{|\mathcal{Q}|}.
\end{aligned}
\label{eq:osc_selection}
\end{equation}

To ensure reproducible selection, ties within each combination are
resolved by higher raw compatibility. Voting ties are resolved
successively by the mean ordinal score across all three criteria,
the median ordinal score, and raw compatibility. Any remaining ties
at either stage favor the lexicographically larger stored
candidate-mask path. Omitting compatibility therefore removes it
only from the ordinal mean; raw compatibility remains a tie-breaker. The quantity $s_{\hat{k}}$ measures selection stability under changes
in the aggregated criteria within the current retained set. It is
not interpreted as a probability of segmentation correctness.
In particular, a singleton retained set has unit stability regardless
of localization quality. OSC outputs the selected mask $m_{\hat{k}}$
and its stability $s_{\hat{k}}$ for subsequent soft pseudo-label
and supervision-weight construction.

\subsection{Reliability Estimation and Pseudo-Label Construction}
\label{subsec:pseudo_label_confidence}

OSC identifies a candidate mask and its selection stability
$s_{\hat{k}}$, but reliability may vary across its pixels, particularly
near region boundaries and locations where candidates disagree.
To modulate their contributions to student training, we construct
a dense soft target $y$ from the selected candidate and estimate
pixel-level weights $w$ using selection stability, local candidate
agreement, and predictive certainty. Encoded supervision targets
take precedence, while teacher predictions supply targets elsewhere.

To match the student training resolution, we align the binary
candidate masks, stored target maps, and masks indicating where
targets are assigned to a common supervision grid using
nearest-neighbor interpolation. Target values and assignment masks
jointly reconstruct the partial supervision map $A$, distinguishing
unknown locations from background. Hereafter, $\Omega$ denotes this
supervision grid, and $A$ and $m_k$ denote their aligned versions. To retain the selected candidate's continuous predictions, we recover
its soft prediction from the stored low-resolution SAM logits:
\begin{equation}
\hat{y}=\sigma\!\left(\mathcal{T}(\ell_{\hat{k}})\right),
\label{eq:soft_target_reconstruction}
\end{equation}
where $\ell_{\hat{k}}$ denotes the selected logits.
The transformation $\mathcal{T}$ restores spatial correspondence
through bilinear upsampling, removal of SAM's preprocessing padding,
and resizing to the supervision resolution. The sigmoid function
$\sigma$ converts logits into probabilities. The aligned binary
mask is used as a fallback only when logits are unavailable. To prevent the recovered teacher prediction from overwriting
existing encoded supervision, we define the encoding-assigned region
$\Omega_{\mathrm{ann}}
=\{i\in\Omega\mid A_i\neq\varnothing\}$
and its complement
$\Omega_{\mathrm{un}}=\Omega\setminus\Omega_{\mathrm{ann}}$.
The final target is
\begin{equation}
y_i=
\begin{cases}
A_i,&i\in\Omega_{\mathrm{ann}},\\
\hat{y}_i,&i\in\Omega_{\mathrm{un}}.
\end{cases}
\label{eq:anchor_preserving_target}
\end{equation}
The labels in $\Omega_{\mathrm{ann}}$ retain the meanings and
assumptions specified by the encoding rules; the selected candidate
provides soft supervision at the remaining locations. Within $\Omega_{\mathrm{un}}$, we first measure pixel-level agreement
among the retained candidates. Let
$K_s=|\mathcal{K}_{\mathrm{safe}}|$. We define
\begin{equation}
\mu_i=\frac{1}{K_s}
\sum_{k\in\mathcal{K}_{\mathrm{safe}}}m_{k,i},
\qquad
g_i=|2\mu_i-1|.
\label{eq:pixel_candidate_agreement}
\end{equation}
where $\mu_i$ is the fraction of retained candidates predicting
foreground at pixel $i$. The score $g_i$ equals one when all
candidates agree and zero when foreground and background decisions
are evenly split. It measures local consensus within the retained
collection, rather than the fraction supporting the selected
candidate's decision at that pixel.

Candidate agreement can coexist with ambiguous probability
predictions. We therefore additionally measure predictive certainty
using normalized binary entropy:
\begin{equation}
\begin{aligned}
h_i
&=1-\frac{\mathcal{H}_{\mathrm{bin}}(y_i)}{\log 2},\\
\mathcal{H}_{\mathrm{bin}}(p)
&=-p\log p-(1-p)\log(1-p).
\end{aligned}
\label{eq:posterior_certainty}
\end{equation}
where $\log$ denotes the natural logarithm. Probabilities are clipped
to $[\epsilon,1-\epsilon]$, with $\epsilon=10^{-6}$, to avoid
evaluating logarithms at zero. The score $h_i$ is low for predictions
near $0.5$ and high for predictions near zero or one. Finally, we combine candidate-level selection stability with the
two pixel-level indicators through a geometric mean:
\begin{equation}
w_i=
\begin{cases}
1,&i\in\Omega_{\mathrm{ann}},\\[3pt]
\left(s_{\hat{k}}g_i h_i\right)^{1/3},
&i\in\Omega_{\mathrm{un}}.
\end{cases}
\label{eq:teacher_weights}
\end{equation}
Unit weights in $\Omega_{\mathrm{ann}}$ prevent candidate disagreement
or teacher uncertainty from attenuating encoded supervision.
Elsewhere, the three indicators jointly determine supervision
strength, with a zero value in any indicator yielding zero
pseudo-label weight. The geometric mean introduces no additional
tunable fusion coefficients. These weights modulate supervision
contributions and are not interpreted as probabilities of pixel-level
correctness. Authentic images receive all-background targets and
unit weights without candidate-agreement computation. We store the soft targets and pixel-level weights $(y,w)$,
while retaining the encoded target maps and their assignment
masks separately.

\subsection{Confidence-Aware Student Optimization}
\label{subsec:student_distillation}

We adopt SCAF~\cite{li2026beyond} as the student model, retaining
its image and foreground/background map inputs. Teacher targets $y$
and reliability weights $w$ are constructed offline and are not
updated from student predictions during training. Supervision sources
differ in spatial coverage: averaging them jointly over pixels may
dilute sparse supervision within dense pseudo-labels. We therefore
prioritize direct targets at designated locations, learn from
confidence-weighted teacher targets elsewhere, and normalize the
two supervision terms separately.

Let $d_i\in[0,1]$ and $v_i\in[0,1]$ denote the direct supervision
target and its weight, respectively. The latter specifies where
the direct target applies and its supervision strength.
For annotation-derived supervision, these quantities are obtained
from the stored encoded target maps and assignment masks.
Images and all supervision maps undergo consistent spatial
transformations, with interpolation allowing fractional targets
and weights. Positive image-level labels alone provide no direct
pixel supervision. In configurations using model-generated auxiliary
targets, their target and weight maps are supplied through the same
interface. These targets constitute additional pseudo-supervision
and are not treated as original annotation facts in safety filtering.

We define $b_i=\mathbf{1}[v_i>0]$ to indicate whether a direct target
applies and set $u_i=(1-b_i)w_i$ as the effective teacher
pseudo-label weight. The direct-supervision and pseudo-label BCE
terms thus operate on disjoint pixel regions, preventing a location
from receiving two BCE targets simultaneously.
Let $p_i=\sigma(z_i)$ denote the student's predicted manipulation
probability, where $z_i$ is its prediction logit. To alleviate class imbalance caused by sparse foreground, we adjust
the positive-class penalty according to the weighted foreground
and background target masses. Given targets $t$ and nonnegative
weights $q$, we define the weighted binary cross-entropy as
\begin{equation}
\begin{aligned}
\gamma(t,q)
&=\operatorname{clip}\!\left(
\frac{\sum_i q_i(1-t_i)}
{\max\!\left(\sum_i q_it_i,\epsilon\right)},
1,\gamma_{\max}
\right),\\
\ell_{\gamma}(p_i,t_i)
&=-\gamma t_i\log p_i-(1-t_i)\log(1-p_i),\\
\mathcal{L}_{\mathrm{BCE}}(p,t,q)
&=
\frac{\sum_i q_i\ell_{\gamma(t,q)}(p_i,t_i)}
{\max\!\left(\sum_i q_i,\epsilon\right)}.
\end{aligned}
\label{eq:balanced_weighted_bce}
\end{equation}
All sums above cover the pixels in the entire mini-batch.
The operator $\operatorname{clip}$ restricts the positive-class
coefficient to $[1,\gamma_{\max}]$, with $\gamma_{\max}=8$ and
$\epsilon=10^{-6}$. In implementation, BCE is evaluated directly
from logits for numerical stability.
The teacher pseudo-label and direct-supervision losses are
$\mathcal{L}_{\mathrm{tea}}
=\mathcal{L}_{\mathrm{BCE}}(p,y,u)$ and
$\mathcal{L}_{\mathrm{dir}}
=\mathcal{L}_{\mathrm{BCE}}(p,d,v)$, respectively.
Each term computes its own positive-class coefficient and is
normalized by its total effective supervision weight; an empty
support contributes zero loss. Separate normalization prevents
direct supervision from being diluted solely because teacher
pseudo-labels cover more pixels. To complement pixel-level BCE with region-overlap supervision,
we use a weighted soft Dice loss~\cite{sudre2017generalised}. Its targets and weights follow
the same direct-supervision priority:
\begin{equation}
\bar{y}_i=b_id_i+(1-b_i)y_i,
\qquad
\bar{w}_i=b_iv_i+(1-b_i)w_i.
\label{eq:mixed_student_supervision}
\end{equation}
For a mini-batch of $N_b$ images, let $\Omega_n$ denote the pixel
domain of image $n$. Dice loss is computed per image and then
averaged over the mini-batch:
\begin{equation}
\mathcal{L}_{\mathrm{dice}}
=
\frac{1}{N_b}\sum_{n=1}^{N_b}
\left[
1-
\frac{
2\sum_{i\in\Omega_n}\bar{w}_ip_i\bar{y}_i+\delta
}{
\sum_{i\in\Omega_n}\bar{w}_i(p_i+\bar{y}_i)+\delta
}
\right],
\label{eq:weighted_soft_dice}
\end{equation}
where $\delta=1$ is a smoothing constant. Teacher reliability
weights therefore modulate both BCE and region-overlap supervision
outside the direct-supervision regions. Each prediction output is optimized with the three loss terms.
The same objective is applied to the main output and two auxiliary
outputs, providing supervision at multiple output levels:
\begin{align}
\mathcal{L}_{\mathrm{distill}}
&=
\mathcal{L}_{\mathrm{tea}}
+\lambda_{\mathrm{dir}}\mathcal{L}_{\mathrm{dir}}
+\lambda_{\mathrm{dice}}\mathcal{L}_{\mathrm{dice}},
\label{eq:student_distillation_loss}\\
\mathcal{L}
&=
\mathcal{L}_{\mathrm{distill}}^{\mathrm{main}}
+\frac{\lambda_{\mathrm{aux}}}{2}
\sum_{j=1}^{2}
\mathcal{L}_{\mathrm{distill}}^{\mathrm{aux},j}.
\label{eq:overall_loss}
\end{align}
The nonnegative coefficients $\lambda_{\mathrm{dir}}$,
$\lambda_{\mathrm{dice}}$, and $\lambda_{\mathrm{aux}}$ control
the contributions of direct supervision, region-overlap supervision,
and auxiliary outputs, respectively.
During inference, only the main prediction output is used,
without accessing training supervision targets or reliability weights
or performing OSC candidate selection.

\begin{table}[!t]
\centering
\caption{The datasets used in our experiments. CM, SP, and IP indicate three common image manipulation types: copy-move, splicing, and inpainting.}
\label{tab:datasets}
\resizebox{0.47\textwidth}{!}{%
\begin{tabular}{@{}ccccccc@{}}
\toprule
\textbf{Dataset} & \textbf{Nums} & \textbf{\#CM} & \textbf{\#SP} & \textbf{\#IP} & \textbf{Train} & \textbf{Test} \\ 
\midrule
CASIAv2~\cite{dong2013casia} & 5123 & 3295 & 1828 & 0 & 5123 & 0 \\ 
Coverage~\cite{wen2016coverage} & 100 & 100 & 0 & 0 & 70 & 30 \\ 
NIST16-C~\cite{ma2024imdlbenco} & 564 & 68 & 288 & 208 & 383 & 181 \\ 
CASIAv1~\cite{dong2013casia} & 920 & 459 & 461 & 0 & 0 & 920 \\
Columbia~\cite{hsu2006columbia} & 180 & 0 & 180 & 0 & 0 & 180 \\ 
Korus~\cite{korus2016evaluation} & 220 & -- & -- & -- & 0 & 220 \\
DSO~\cite{de2013exposing} & 100 & 0 & 100 & 0 & 0 & 100 \\
IMD2020~\cite{Novozamsky_2020_WACV} & 2010 & -- & -- & -- & 0 & 2010 \\
CocoGlide~\cite{nichol2021glide} & 512 & -- & --& -- & 0 &512 \\
ITW~\cite{huh2018fighting} & 201 & 0 & 201 & 0 &0 &201\\
\bottomrule
\end{tabular}%
}
\end{table}

\begin{table*}[!t]
\centering
\caption{Performance comparison with other advanced IML methods in terms of F1 score (fixed threshold at 0.5). The best results are highlighted in bold, and the second-best results are underlined.}
\label{tab:main_comparison}
\resizebox{\textwidth}{!}{
\setlength{\tabcolsep}{2.6pt}
\renewcommand{\arraystretch}{1.02}
\begin{tabular}{@{}lccc*{11}{c}@{}}
\toprule
\multirow{2}{*}{\textbf{Method}} & \multirow{2}{*}{\textbf{Pub.}} & \multirow{2}{*}{\textbf{Sup.}} & \multirow{2}{*}{\textbf{Epoch}}
& \multicolumn{4}{c}{\textbf{ID $F_1$}}
& \multicolumn{7}{c}{\textbf{OOD $F_1$}} \\
\cmidrule(lr){5-8} \cmidrule(lr){9-15}
& & & & \textbf{CASIAv1} & \textbf{Coverage} & \textbf{NIST16-C} & \textbf{Avg.}
& \textbf{Columbia} & \textbf{IMD2020} & \textbf{Korus} & \textbf{ITW} & \textbf{CocoGlide} & \textbf{DSO} & \textbf{Avg.} \\
\midrule

\multirow{3}{*}{MFI-Net} & \multirow{3}{*}{TCSVT'24} & \multirow{3}{*}{Full} & 10
& 0.604 & 0.521 & 0.426 & 0.517
& 0.652 & 0.373 & 0.241 & 0.315 & 0.323 & 0.188 & 0.349 \\

 & & & 20
& 0.571 & 0.448 & 0.432 & 0.483
& 0.342 & 0.331 & 0.176 & 0.255 & 0.204 & 0.081 & 0.232 \\

 & & & 70
& 0.669 & 0.478 & 0.451 & 0.533
& 0.544 & 0.288 & 0.166 & 0.236 & 0.224 & 0.094 & 0.259 \\

\midrule

\multirow{3}{*}{Mesorch} & \multirow{3}{*}{AAAI'25} & \multirow{3}{*}{Full} & 10
& 0.611 & 0.307 & 0.398 & 0.439
& 0.537 & 0.349 & 0.176 & 0.308 & 0.175 & 0.190 & 0.289 \\

 & & & 20
& 0.649 & 0.449 & 0.440 & 0.512
& 0.655 & 0.364 & 0.185 & 0.294 & 0.159 & 0.196 & 0.309 \\

 & & & 70
& 0.692 & 0.613 & 0.511 & 0.605
& 0.598 & 0.348 & 0.174 & 0.283 & 0.224 & 0.221 & 0.308 \\

\midrule

\multirow{3}{*}{SparseViT} & \multirow{3}{*}{AAAI'25} & \multirow{3}{*}{Full} & 10
& 0.466 & 0.250 & 0.342 & 0.353
& 0.378 & 0.272 & 0.146 & 0.237 & 0.243 & 0.158 & 0.239 \\

 & & & 20
& 0.517 & 0.339 & 0.356 & 0.404
& 0.380 & 0.283 & 0.141 & 0.236 & 0.180 & 0.129 & 0.225 \\

 & & & 70
& 0.526 & 0.660 & 0.493 & 0.559
& 0.543 & 0.339 & 0.160 & 0.276 & 0.220 & 0.192 & 0.288 \\

\midrule

\multirow{3}{*}{PIM-Net} & \multirow{3}{*}{PR'25} & \multirow{3}{*}{Full} & 10
& 0.568 & 0.536 & 0.444 & 0.516
& 0.636 & 0.358 & 0.197 & 0.301 & 0.282 & 0.178 & 0.325 \\

 & & & 20
& 0.587 & 0.655 & 0.496 & 0.579
& 0.669 & 0.372 & 0.237 & 0.314 & 0.324 & 0.117 & 0.339 \\

 & & & 70
& 0.658 & 0.684 & \underline{0.596} & 0.646
& 0.761 & 0.413 & 0.228 & 0.333 & 0.354 & 0.282 & 0.395 \\

\midrule

\multirow{2}{*}{PIM} & \multirow{2}{*}{TPAMI'25} & \multirow{2}{*}{Full} & 10
& 0.548 & 0.583 & 0.330 & 0.487
& 0.597 & 0.349 & 0.171 & 0.325 & 0.394 & 0.175 & 0.335 \\

 & & & 20
& 0.577 & 0.555 & 0.411 & 0.514
& 0.626 & 0.345 & 0.204 & 0.332 & 0.269 & 0.165 & 0.324 \\

\midrule

\multirow{3}{*}{PNF-Net} & \multirow{3}{*}{TCSVT'26} & \multirow{3}{*}{Full} & 10
& 0.673 & 0.565 & 0.526 & 0.588
& 0.787 & 0.455 & 0.252 & 0.387 & 0.318 & 0.274 & 0.412 \\

 & & & 20
& 0.694 & 0.649 & 0.557 & 0.633
& 0.757 & 0.454 & 0.241 & 0.377 & 0.366 & 0.334 & 0.422 \\

 & & & 70
& \textbf{0.753} & 0.746 & \textbf{0.599} & \textbf{0.699}
& 0.726 & 0.437 & 0.214 & 0.362 & 0.328 & 0.215 & 0.380 \\

\midrule

\multirow{3}{*}{Co-Transformers} & \multirow{3}{*}{AAAI'26} & \multirow{3}{*}{Full} & 10
& 0.646 & 0.445 & 0.359 & 0.483
& 0.798 & 0.323 & 0.092 & 0.290 & 0.181 & 0.127 & 0.302 \\

 & & & 20
& 0.667 & 0.606 & 0.466 & 0.580
& 0.867 & 0.359 & 0.155 & 0.289 & 0.268 & 0.174 & 0.352 \\

 & & & 70
& 0.709 & \underline{0.760} & 0.506 & 0.658
& 0.867 & 0.414 & 0.155 & 0.362 & 0.273 & 0.147 & 0.370 \\

\midrule

\multirow{3}{*}{FRD-Net} & \multirow{3}{*}{AAAI'26} & \multirow{3}{*}{Full} & 10
& 0.685 & 0.485 & 0.445 & 0.539
& 0.878 & 0.426 & 0.214 & 0.394 & 0.294 & 0.181 & 0.398 \\

 & & & 20
& 0.686 & 0.595 & 0.472 & 0.584
& \underline{0.883} & 0.436 & 0.211 & 0.376 & 0.295 & 0.239 & 0.407 \\

 & & & 70
& 0.722 & \textbf{0.784} & 0.537 & \underline{0.681}
& \textbf{0.895} & 0.488 & 0.263 & 0.437 & 0.377 & 0.334 & \textbf{0.465} \\

\midrule

\multirow{3}{*}{RelayFormer} & \multirow{3}{*}{ICLR'26} & \multirow{3}{*}{Full} & 10
& 0.476 & 0.228 & 0.284 & 0.329
& 0.440 & 0.254 & 0.105 & 0.219 & 0.229 & 0.070 & 0.220 \\

 & & & 20
& 0.567 & 0.356 & 0.370 & 0.431
& 0.508 & 0.313 & 0.159 & 0.277 & 0.240 & 0.113 & 0.268 \\

 & & & 70
& 0.683 & 0.508 & 0.527 & 0.573
& 0.662 & 0.399 & 0.238 & 0.363 & 0.276 & 0.239 & 0.363 \\

\midrule

\multirow{3}{*}{SCAF} & \multirow{3}{*}{AAAI'26} & \multirow{3}{*}{Scribble} & 10
& 0.379 & 0.375 & 0.168 & 0.307
& 0.357 & 0.314 & 0.193 & 0.283 & 0.344 & 0.261 & 0.292 \\

 & & & 20
& 0.672 & 0.687 & 0.368 & 0.576
& 0.691 & 0.516 & 0.243 & \textbf{0.489} & \textbf{0.536} & 0.237 & 0.452 \\

 & & & 70
& 0.745 & 0.694 & 0.239 & 0.559
& 0.646 & 0.439 & 0.181 & 0.392 & 0.479 & 0.206 & 0.390 \\

\midrule

\multirow{2}{*}{BoxPromptIML} & \multirow{2}{*}{AAAI'26} & \multirow{2}{*}{Box} & 10
& 0.529 & 0.414 & 0.375 & 0.439
& 0.335 & 0.323 & 0.241 & 0.302 & 0.216 & 0.241 & 0.276 \\

 & & & 20
& 0.543 & 0.418 & 0.409 & 0.457
& 0.378 & 0.321 & 0.248 & 0.305 & 0.237 & 0.253 & 0.290 \\

\midrule

\multirow{9}{*}{Ours} & \multirow{9}{*}{--} & \multirow{3}{*}{Scribble} & 10
& 0.558 & 0.490 & 0.379 & 0.476
& 0.614 & 0.421 & 0.232 & 0.423 & 0.455 & 0.336 & 0.414 \\

 & & & 20
& 0.709 & 0.706 & 0.497 & 0.637
& 0.726 & \underline{0.518} & \underline{0.276} & 0.472 & 0.460 & 0.310 & \underline{0.460} \\

 & & & 70
& \underline{0.749} & 0.685 & 0.546 & 0.660
& 0.694 & \textbf{0.539} & \textbf{0.291} & \underline{0.485} & \underline{0.485} & 0.214 & 0.451 \\

\cmidrule(lr){3-15}

 & & \multirow{3}{*}{Box} & 10
& 0.489 & 0.464 & 0.361 & 0.438
& 0.516 & 0.406 & 0.231 & 0.444 & 0.426 & \underline{0.361} & 0.397 \\

 & & & 20
& 0.644 & 0.634 & 0.446 & 0.575
& 0.483 & 0.509 & 0.264 & 0.458 & 0.446 & \textbf{0.397} & 0.426 \\

 & & & 70
& 0.659 & 0.598 & 0.506 & 0.588
& 0.473 & 0.489 & 0.236 & 0.405 & 0.412 & 0.331 & 0.391 \\

\cmidrule(lr){3-15}

 & & \multirow{3}{*}{Point} & 10
& 0.514 & 0.401 & 0.231 & 0.382
& 0.425 & 0.347 & 0.180 & 0.324 & 0.304 & 0.251 & 0.305 \\

 & & & 20
& 0.546 & 0.399 & 0.277 & 0.407
& 0.414 & 0.376 & 0.201 & 0.349 & 0.328 & 0.240 & 0.318 \\

 & & & 70
& 0.314 & 0.214 & 0.353 & 0.294
& 0.161 & 0.252 & 0.136 & 0.116 & 0.099 & 0.095 & 0.143 \\

\bottomrule
\end{tabular}
}
% \vspace{2pt}

\end{table*}

\begin{table}[t]
\centering
\caption{Comparison with image-level weakly supervised IML methods.}
\label{tab:image_level}
\setlength{\tabcolsep}{2pt}
\renewcommand{\arraystretch}{1.08}
\resizebox{\columnwidth}{!}{%
\begin{tabular}{@{}lccccccc@{}}
\toprule
\textbf{Method} & \textbf{Epoch} & \textbf{CASIAv1} & \textbf{Coverage} & \textbf{NIST16} & \textbf{IMD2020} & \textbf{Columbia} & \textbf{Avg.} \\
\midrule
WSCL~\cite{zhai2023towards} & 60 & 0.153 & 0.201 & 0.099 & 0.173 & 0.362 & 0.198 \\
EdgeCAM~\cite{ZHOU2024123501} & 100 & 0.301 & 0.262 & 0.254 & 0.242 & 0.470 & 0.306 \\
SOWCL~\cite{10889843} & -- & 0.334 & 0.239 & \textbf{0.288} & 0.259 & 0.385 & 0.301 \\
WSCCL~\cite{bai2025weakly} & 100 & 0.349 & 0.281 & \underline{0.278} & 0.259 & \underline{0.516} & \underline{0.337} \\
\midrule
\multirow{3}{*}{Ours} & 10 & 0.279 & 0.307 & 0.176 & 0.232 & \textbf{0.517} & 0.302 \\
 & 20 & \underline{0.365} & \underline{0.395} & 0.192 & \underline{0.264} & 0.474 & \textbf{0.338} \\
 & 70 & \textbf{0.382} & \textbf{0.411} & 0.184 & \textbf{0.288} & 0.380 & 0.329 \\
\bottomrule
\end{tabular}%
}
% \vspace{2pt}
\end{table}

\section{Experiments}
\label{sec:experiments}

\subsection{Datasets and Evaluation Metrics}

We conduct experiments on 10 widely used IML benchmarks: CASIAv2~\cite{dong2013casia}, Coverage~\cite{wen2016coverage}, NIST16-C~\cite{ma2024imdlbenco}, CASIAv1~\cite{dong2013casia}, Columbia~\cite{hsu2006columbia}, ITW~\cite{huh2018fighting}, DSO~\cite{de2013exposing}, Korus~\cite{korus2016evaluation}, IMD2020~\cite{Novozamsky_2020_WACV}, and CocoGlide~\cite{nichol2021glide}. Following Li et al.~\cite{li2026prototype}, we adopt the same training data split. The remaining data and corresponding test splits are used to evaluate in-distribution (ID) performance and out-of-distribution (OOD) generalization. For ID evaluation, CASIAv1 covers copy-move and splicing, Coverage includes challenging copy-move cases with similar source and target regions, and NIST16-C addresses the label-leakage issue in the original NIST16~\cite{guan2019mfc} while covering complex manipulations. For OOD evaluation, Columbia focuses on splicing boundaries, ITW and DSO cover object-level manipulations and realistic scenarios, while Korus, IMD2020, and CocoGlide introduce diverse real-world scenes and generative manipulations. Dataset statistics, manipulation types, and splits are summarized in Table~\ref{tab:datasets}.

We evaluate localization performance using the pixel-level $F_1$ score, the harmonic mean of precision and recall~\cite{ma2024imdlbenco}. As noted by Ma et al.~\cite{ma2024imdlbenco}, AUC can overestimate localization performance due to the severe imbalance between manipulated and authentic pixels. We therefore use $F_1$ as the sole evaluation metric, with a fixed binarization threshold of 0.5 for fair comparison across methods.

\subsection{Implementation Details}
We adopt PVTv2~\cite{wang2022pvt} pretrained on ImageNet as the backbone of the student localization network. Compared with the conference version~\cite{li2026beyond}, the student network is trained through a unified supervision interface that supports scribble, box, point, and mixed weak annotations. All input images and supervision maps are resized to $416\times416$. SAM-based candidate generation and OSC-based pseudo-label selection are performed offline prior to training the student model. We train the model on two NVIDIA RTX 3090 GPUs using the AdamW optimizer~\cite{loshchilovdecoupled} with a batch size of 32 and an initial learning rate of 1e-4.

\begin{figure*}[!t]
\centering
\includegraphics[width=\textwidth]{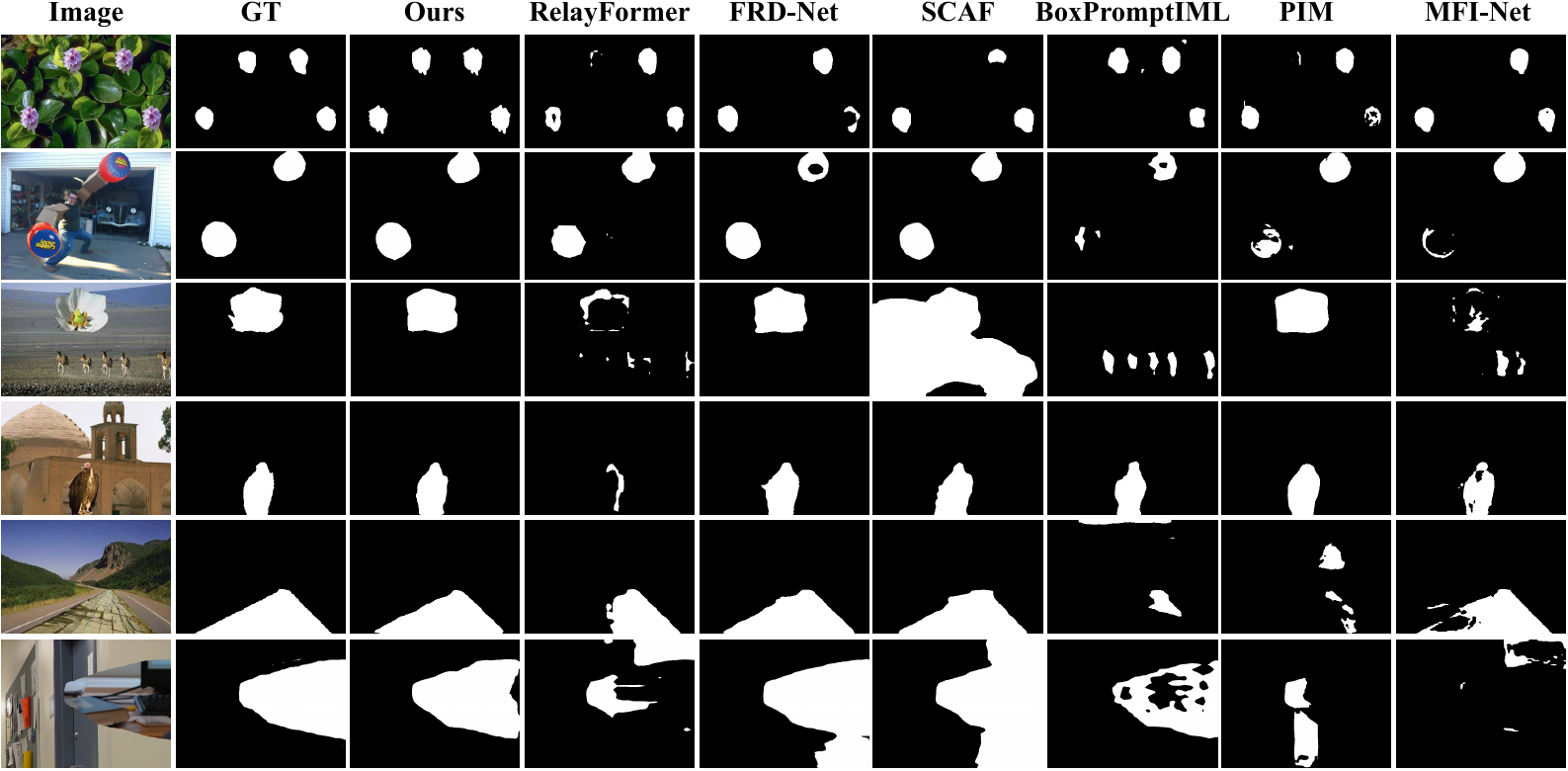}
\caption{
Qualitative comparison of image manipulation localization on six representative examples.
From left to right: input image, ground truth (GT), UniIML (Ours), RelayFormer, FRD-Net, SCAF, BoxPromptIML, PIM, and MFI-Net.
The results of UniIML are obtained under scribble supervision, and all methods are visualized using models trained for 20 epochs.
White and black denote manipulated and authentic regions, respectively.
}
\label{fig:qualitative_comparison}
\end{figure*}

\subsection{Comparison With State-of-the-Art Methods}
\subsubsection{Baseline Models}

We compare UniIML with fifteen representative image manipulation localization (IML) methods, covering both fully supervised and weakly supervised settings. The fully supervised baselines include MFI-Net~\cite{mfi2024}, Mesorch~\cite{zhu2025mesoscopic}, SparseViT~\cite{su2025sparsevit}, PIM-Net~\cite{bai2025pimnet}, PIM~\cite{kong2025pixel}, PNF-Net~\cite{li2026prototype}, Co-Transformers~\cite{zhang2026collaborative}, FRD-Net~\cite{chen2026amplifying}, and RelayFormer~\cite{huang2026relayformer}. The weakly supervised baselines include SCAF~\cite{li2026beyond} with scribble supervision, BoxPromptIML~\cite{guo2026passive} with box supervision, and the image-level methods WSCL~\cite{zhai2023towards}, EdgeCAM~\cite{ZHOU2024123501}, SOWCL~\cite{10889843}, and WSCCL~\cite{bai2025weakly}.

Since reproducible implementations of the aforementioned image-level supervised methods are not publicly available, we report their results directly from the original publications. To ensure a fair comparison, we adopt the same training benchmark for UniIML under image-level supervision, using only authentic and manipulated images from CASIAv2~\cite{dong2013casia} and their image-level labels. All remaining baselines are reimplemented following the experimental benchmark of Li et al.~\cite{li2026prototype} in a unified experimental environment to ensure a consistent comparison. Furthermore, to the best of our knowledge, no prior work has investigated point-supervised IML. Therefore, no directly comparable baseline is available in this setting.

\subsubsection{Image Manipulation Localization} 
Table~\ref{tab:main_comparison} compares UniIML under scribble, bounding-box, and point supervision, and further benchmarks it against existing weakly supervised and fully supervised IML methods. We first examine the effect of different weak supervision forms. A consistent performance hierarchy can be observed, with scribble supervision generally outperforming bounding-box supervision, while point supervision yields comparatively lower performance. This trend is closely related to the amount and reliability of spatial information provided by each annotation type. Scribble annotations offer verified foreground and background pixels distributed over the image, thereby providing relatively informative constraints on both the location and extent of manipulated regions. Bounding boxes effectively restrict the feasible support of the manipulation, but still leave substantial ambiguity within the box interior. In contrast, point annotations provide only sparse positional cues and contain little direct information about region extent, shape, or boundary. As a result, the candidate space becomes increasingly ambiguous as the supervision becomes weaker. These results indicate that, although UniIML can accommodate heterogeneous annotation forms within a unified framework, the attainable performance still depends on the spatial informativeness of the available supervision.

An interesting observation is that the performance under point supervision degrades substantially when the training schedule is extended to 70 epochs. We attribute this behavior to the extreme sparsity of point-level constraints. Since only a few isolated coordinates are available to validate candidate masks, multiple spatially distinct candidates may satisfy the same point annotations, making erroneous pseudo-labels more difficult to identify and reject. During prolonged training, such noisy pseudo-labels can be repeatedly reinforced by the model and subsequently reused in later optimization stages, leading to progressive error accumulation and overfitting to inaccurate spatial hypotheses. In contrast, scribble and box annotations provide stronger spatial constraints that more effectively restrict the feasible candidate space and reduce this ambiguity. This observation suggests that, under extremely sparse supervision, extending the training duration does not necessarily improve IML accuracy and may instead amplify pseudo-label noise.

Under the same supervision setting, UniIML consistently compares favorably with existing weakly supervised approaches. It outperforms SCAF~\cite{li2026beyond} under scribble supervision and BoxPromptIML~\cite{guo2026passive} under bounding-box supervision. We attribute these gains primarily to how UniIML exploits weak annotations. Rather than directly expanding sparse or coarse annotations into dense supervision, UniIML first converts them into foreground and background constraints and then leverages SAM to generate candidate masks with coherent region structures. OSC subsequently evaluates the consistency between these candidates and the available weak constraints, retaining only those that are compatible with the verified evidence. In this way, weak annotations are used as constraints for validating candidate regions instead of being naively propagated into pixel-level labels, which helps suppress erroneous spatial expansion and reduces the noise introduced by unreliable pseudo-labels. This design is particularly beneficial when the available supervision is sparse or structurally incomplete.

\begin{figure*}[!t]
\centering
\subfloat[In-Distribution Datasets.\label{fig:heterogeneous_ind}]{%
\includegraphics[width=0.48\textwidth]{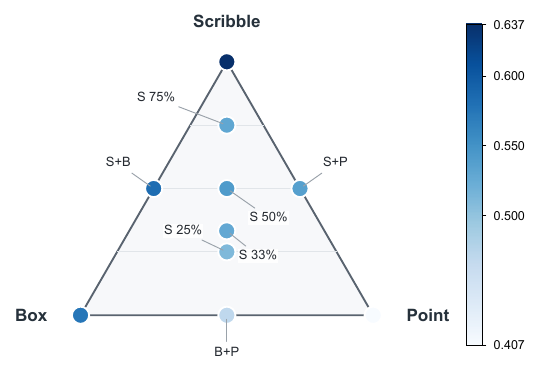}}\hfill
\subfloat[Out-Of-Distribution Datasets.\label{fig:heterogeneous_ood}]{%
\includegraphics[width=0.48\textwidth]{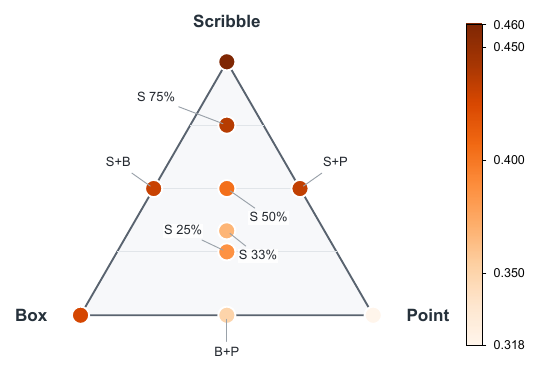}}
\caption{Localization performance under varying compositions of mixed weak annotations on (a) in-distribution (ID) and (b) out-of-distribution (OOD) datasets. The three vertices correspond to single-type supervision, the edges denote pairwise mixtures, and the interior points represent three-way mixtures with different annotation ratios. The values shown in (a) and (b) denote the mean F1 scores across the ID and OOD datasets, respectively.}
\label{fig:heterogeneous_composition}
\end{figure*}

\begin{table}[!t]
\centering
\caption{Ablation results under image-level supervision.}
\label{tab:ablation_image}
\setlength{\tabcolsep}{2pt}
\scriptsize
\resizebox{\columnwidth}{!}{%
\begin{tabular}{@{}lcccccc@{}}
\toprule
\textbf{Settings} & \textbf{CASIAv1} & \textbf{Coverage} & \textbf{NIST16} & \textbf{IMD2020} & \textbf{Columbia} & \textbf{Avg.} \\
\midrule
w/o OSC & 0.293 & 0.292 & 0.175 & 0.256 & 0.367 & 0.276 \\
% w/o safety filtering & 0.322 & 0.347 & 0.170 & 0.257 & 0.486 & 0.316 \\
% w/o ordinal consensus
% voting & 0.360 & 0.359 & 0.179 & 0.263 & 0.452 & 0.323 \\
\midrule
w/o confidence weighting & 0.296 & 0.329 & 0.155 & 0.233 & 0.507 & 0.304 \\
w/o direct supervision & 0.165 & 0.209 & 0.159 & 0.179 & 0.285 & 0.199 \\
w/o weighted Dice loss & 0.304 & 0.339 & 0.171 & 0.243 & 0.444 & 0.300 \\
\midrule
\textbf{Full UniIML} & \textbf{0.365} & \textbf{0.395} & \textbf{0.192} & \textbf{0.264} & \textbf{0.474} & \textbf{0.338} \\
\bottomrule
\end{tabular}%
}
% \vspace{2pt}
\end{table}

It is also noteworthy that UniIML can match or even surpass several fully supervised methods in some settings, particularly on OOD benchmarks. One possible explanation is that, although fully supervised methods have access to precise pixel-level masks, their optimization can more strongly fit dataset-specific appearance patterns and annotation distributions. In contrast, UniIML constructs supervision through weak constraints, candidate generation, and consistency-based selection, which reduces its reliance on dense dataset-specific masks and encourages the model to exploit more transferable manipulation cues. Consequently, despite using substantially weaker annotations, UniIML can still obtain supervision signals with strong cross-dataset generalization. Overall, these results suggest that the advantage of UniIML does not arise from stronger annotation, but from a more effective and robust utilization of limited supervision, leading to a favorable trade-off among annotation cost, supervision flexibility, and localization performance.

For image-level supervision, Table~\ref{tab:image_level} shows that UniIML remains competitive despite being trained with only binary authenticity labels and without any explicit spatial annotations. In contrast to prior image-level approaches that mainly rely on classification responses or global consistency cues, UniIML further converts coarse image-level evidence into foreground and background constraints, which are then used to guide candidate mask generation and selection. This design enables the framework to recover more informative spatial supervision from weak image-level labels.

Overall, UniIML achieves the best average performance among the compared methods at 20 epochs, while maintaining strong results on several individual benchmarks. With longer training, the performance on CASIAv1, Coverage, and IMD2020 further improves, whereas the score on Columbia decreases, suggesting that additional optimization does not necessarily translate into better cross-dataset generalization under such weak supervision. This observation is consistent with the limited spatial information provided by image-level labels: once the model begins to overfit dataset-specific appearance cues, the quality of the inferred localization supervision may deteriorate. Therefore, the results indicate that UniIML can effectively extract useful localization cues from image-level annotations, while also highlighting the importance of controlling pseudo-label noise and overfitting when spatial supervision is entirely absent.

\begin{table*}[!t]
\centering
\caption{Ablation results under scribble, point, and bounding-box supervision.}
\label{tab:ablation_spatial}
\setlength{\tabcolsep}{2pt}
\scriptsize
\resizebox{\textwidth}{!}{%
\begin{tabular}{@{}llccccccccccc@{}}
\toprule
\textbf{Supervision} & \textbf{Settings} &
\textbf{CASIAv1} & \textbf{Coverage} & \textbf{NIST16-C} &
\textbf{ID Avg.} & \textbf{Columbia} & \textbf{IMD2020} &
\textbf{Korus} & \textbf{ITW} & \textbf{CocoGlide} &
\textbf{DSO} & \textbf{OOD Avg.} \\
\midrule

% ==================== Scribble ====================
\multirow{7}{*}{Scribble}
& w/o OSC
& 0.628 & 0.611 & 0.379 & 0.539
& 0.621 & 0.460 & 0.243 & 0.417 & 0.441 & 0.314 & 0.416 \\

& w/o safety filtering
& 0.653 & 0.599 & 0.418 & 0.557
& 0.625 & 0.459 & 0.245 & 0.469 & 0.492 & 0.353 & 0.440 \\

& w/o ordinal consensus voting
& 0.646 & 0.635 & 0.435 & 0.572
& 0.650 & 0.473 & 0.264 & 0.435 & 0.465 & 0.319 & 0.434 \\

\cmidrule(lr){2-13}

& w/o confidence weighting
& 0.613 & 0.619 & 0.404 & 0.545
& 0.643 & 0.435 & 0.257 & 0.442 & 0.474 & 0.349 & 0.434 \\

& w/o direct supervision
& 0.655 & 0.629 & 0.439 & 0.574
& 0.613 & 0.469 & 0.252 & 0.426 & 0.470 & 0.317 & 0.424 \\

& w/o weighted Dice loss
& 0.635 & 0.596 & 0.407 & 0.546
& 0.632 & 0.445 & 0.231 & 0.444 & 0.462 & 0.308 & 0.420 \\

\cmidrule(lr){2-13}

& \textbf{Full UniIML}
& \textbf{0.709} & \textbf{0.706} & \textbf{0.497} & \textbf{0.637}
& \textbf{0.726} & \textbf{0.518} & \textbf{0.276} & \textbf{0.472}
& \textbf{0.460} & \textbf{0.310} & \textbf{0.460} \\

\midrule

% ==================== Point ====================
\multirow{7}{*}{Point}
& w/o OSC
& 0.483 & 0.371 & 0.219 & 0.358
& 0.337 & 0.324 & 0.172 & 0.280 & 0.313 & 0.234 & 0.277 \\

& w/o safety filtering
& 0.520 & 0.409 & 0.261 & 0.397
& 0.406 & 0.331 & 0.191 & 0.300 & 0.309 & 0.229 & 0.294 \\

& w/o ordinal consensus voting
& 0.504 & 0.400 & 0.252 & 0.386
& 0.386 & 0.339 & 0.193 & 0.305 & 0.328 & 0.209 & 0.293 \\

\cmidrule(lr){2-13}

& w/o confidence weighting
& 0.527 & 0.390 & 0.271 & 0.396
& 0.411 & 0.349 & 0.200 & 0.306 & 0.309 & 0.219 & 0.299 \\

& w/o direct supervision
& 0.284 & 0.181 & 0.364 & 0.276
& 0.171 & 0.260 & 0.167 & 0.184 & 0.149 & 0.146 & 0.180 \\

& w/o weighted Dice loss
& 0.525 & 0.365 & 0.250 & 0.380
& 0.353 & 0.340 & 0.180 & 0.308 & 0.283 & 0.207 & 0.278 \\

\cmidrule(lr){2-13}

& \textbf{Full UniIML}
& \textbf{0.546} & \textbf{0.399} & \textbf{0.277} & \textbf{0.407}
& \textbf{0.414} & \textbf{0.376} & \textbf{0.201} & \textbf{0.349}
& \textbf{0.328} & \textbf{0.240} & \textbf{0.318} \\

\midrule

% ==================== Box ====================
\multirow{7}{*}{Box}
& w/o OSC
& 0.495 & 0.499 & 0.381 & 0.458
& 0.463 & 0.406 & 0.214 & 0.395 & 0.403 & 0.336 & 0.370 \\

& w/o safety filtering
& 0.533 & 0.475 & 0.383 & 0.464
& 0.551 & 0.433 & 0.238 & 0.439 & 0.445 & 0.367 & 0.412 \\

& w/o ordinal consensus voting
& 0.531 & 0.507 & 0.388 & 0.475
& 0.572 & 0.423 & 0.227 & 0.442 & 0.441 & 0.345 & 0.408 \\

\cmidrule(lr){2-13}

& w/o confidence weighting
& 0.553 & 0.562 & 0.406 & 0.507
& 0.566 & 0.452 & 0.238 & 0.459 & 0.442 & 0.376 & 0.422 \\

& w/o direct supervision
& 0.653 & 0.583 & 0.450 & 0.562
& 0.565 & 0.459 & 0.238 & 0.441 & 0.464 & 0.350 & 0.419 \\

& w/o weighted Dice loss
& 0.484 & 0.463 & 0.352 & 0.433
& 0.516 & 0.385 & 0.203 & 0.418 & 0.437 & 0.355 & 0.385 \\

\cmidrule(lr){2-13}

& \textbf{Full UniIML}
& \textbf{0.644} & \textbf{0.634} & \textbf{0.446} & \textbf{0.575}
& \textbf{0.483} & \textbf{0.509} & \textbf{0.264} & \textbf{0.458}
& \textbf{0.446} & \textbf{0.397} & \textbf{0.426} \\

\bottomrule
\end{tabular}%
}
% \vspace{2pt}
\end{table*}

\subsubsection{Visual Comparison}
To qualitatively evaluate the IML performance under diverse manipulation scenarios, Fig.~\ref{fig:qualitative_comparison} presents six representative examples, including images with multiple manipulated regions (rows 1--2), small manipulated regions (rows 3--4), and large manipulated regions (rows 5--6). Notably, the results of UniIML (Ours) are obtained under scribble supervision, and all methods are evaluated using models trained for 20 epochs to ensure a consistent training budget. For images containing multiple manipulated regions, UniIML accurately localizes spatially separated targets while preserving their shapes and boundaries. In contrast, several compared methods miss individual manipulated regions or produce incomplete responses. This behavior is associated with the unified constraint interface, where sparse positive anchors provide explicit spatial cues for disconnected manipulated regions and facilitate their simultaneous localization. For small manipulated regions, UniIML retains compact and concentrated responses on the target areas, whereas some competing methods suffer from missed detections, fragmented predictions, or spurious responses in authentic regions. This improvement benefits from OSC, which promotes consistency with the weak constraints and filters candidate masks that conflict with verified authentic regions. For large manipulated regions, UniIML produces spatially continuous masks with more complete coverage and fewer internal holes, while several baselines exhibit fragmentation or under-segmentation. These results demonstrate the effectiveness of consensus-based candidate selection in recovering coherent manipulation regions. Overall, despite relying only on scribble supervision, UniIML consistently produces cleaner and more complete localization results across different manipulation scales and configurations, with fewer false-positive and false-negative responses and better qualitative agreement with the ground truth.

\begin{figure}[!t]
\centering
\includegraphics[width=\columnwidth]{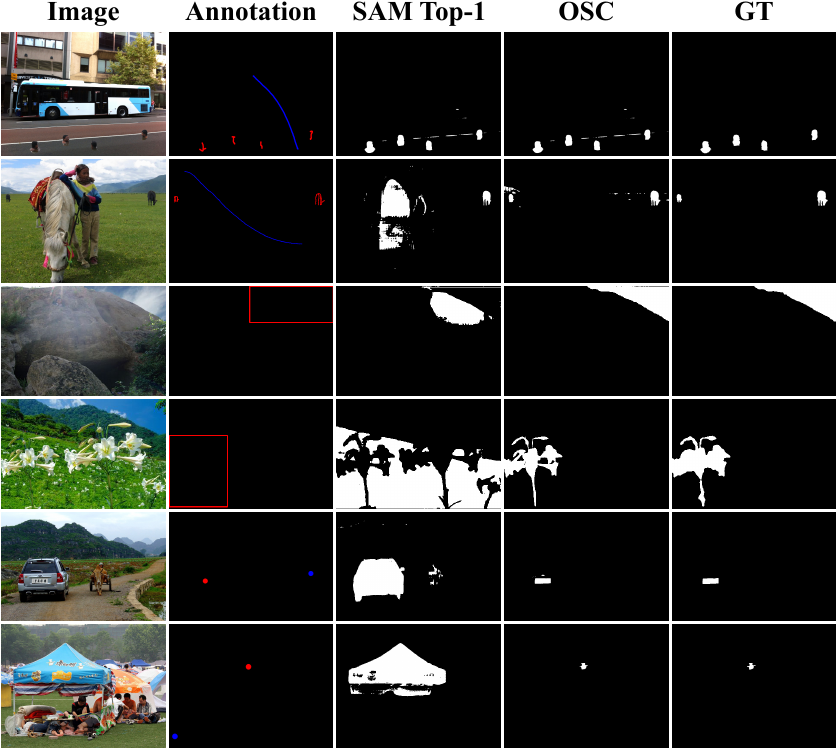}
\caption{Case studies of OSC-based candidate selection under scribble supervision (top two rows), box supervision (middle two rows), and point supervision (bottom two rows). From left to right: input image, weak annotation, SAM Top-1 mask, OSC-selected mask, and ground truth (GT). White indicates manipulated regions, while black indicates authentic regions.}
\label{fig:osc_case_study}
\end{figure}

\subsection{Mixed Weak-Annotation Supervision.}
In the mixed-supervision setting, each manipulated training image is associated with only one annotation type. Different experimental configurations vary only in the proportions of annotation types, while the candidate mask generation and selection procedures, optimization objectives, student backbone, and evaluation protocol remain identical to those used in the single-type supervision experiments. All models are trained using the standard 20-epoch schedule.

\begin{figure}[!t]
\centering
\includegraphics[width=\columnwidth]{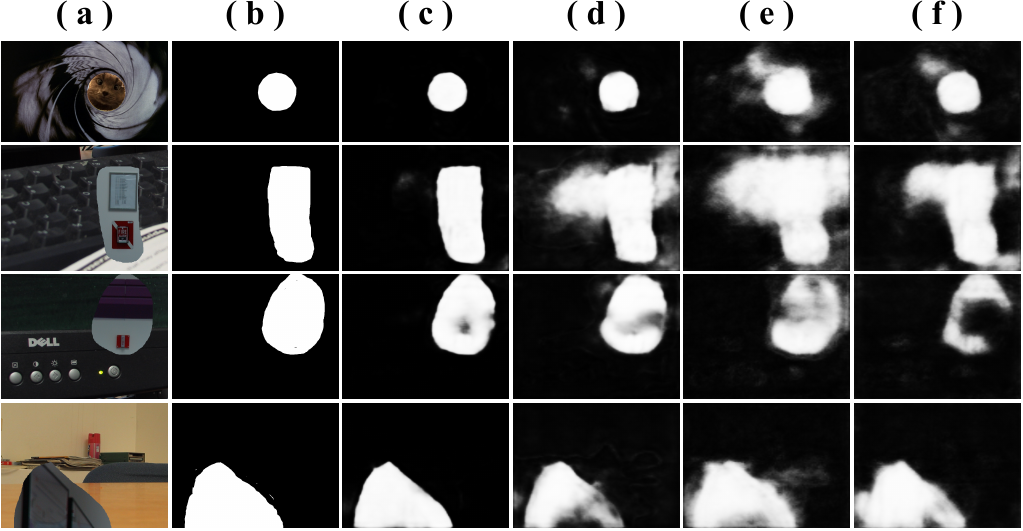}
\caption{Qualitative comparison of student optimization variants under scribble supervision:
(a) input image,
(b) ground truth (GT),
(c) full UniIML (Ours),
(d) without direct supervision,
(e) without confidence weighting, and
(f) without the weighted Dice loss.}
\label{fig:student_ablation_visual}
\end{figure}

Fig.~\ref{fig:heterogeneous_composition} presents the localization performance under single-type supervision, pairwise annotation mixtures, and three-way combinations on the in-distribution (ID) and out-of-distribution (OOD) benchmarks. Under single-type supervision, scribble annotations achieve the highest average performance on both ID and OOD benchmarks, with scores of 0.637 and 0.460, respectively, whereas point supervision yields the lowest scores of 0.407 and 0.318. Among the pairwise configurations, mixtures involving scribble annotations (S+B and S+P) consistently outperform the combination of bounding boxes and points (B+P) on both ID and OOD benchmarks. Under the three-way mixture setting, UniIML maintains effective localization performance across different annotation compositions, although the results vary with the proportions of different annotation types. In particular, when scribble annotations account for 75\% of the training annotations, the model achieves an average OOD performance of 0.436, approaching the 0.460 obtained under scribble-only supervision. However, its corresponding ID performance remains below that of scribble-only supervision. These results suggest that the spatial informativeness of different annotation types plays an important role in localization performance. Specifically, scribble annotations provide distributed and trusted foreground and background cues within manipulated and surrounding regions, offering relatively informative spatial constraints that help identify candidate masks more consistent with the available annotation evidence. In contrast, bounding boxes mainly constrain the coarse spatial extent of manipulated regions, while isolated point annotations provide only sparse positional cues with limited information about region shape and boundaries. Consequently, mixtures involving scribble annotations provide stronger evidence for candidate mask validation, whereas the combination of bounding boxes and points leaves greater ambiguity in the candidate space. This also helps explain why mixtures containing scribble annotations generally achieve better localization performance and why a higher proportion of scribble annotations can yield OOD performance approaching that of scribble-only supervision.

Overall, these results demonstrate that UniIML can effectively utilize different types and proportions of weak annotations within a unified framework while preserving the trusted spatial constraints provided by each annotation type. Through a shared candidate mask generation and selection pipeline, together with a student training strategy that combines direct annotation-derived supervision and reliability-weighted pseudo-label distillation, UniIML enables joint learning from multiple weak annotation types without requiring annotation-specific branches or separate training pipelines. This demonstrates its adaptability to mixed-annotation settings.

\begin{table*}[!t]
\centering
\caption{Robustness to synthetic image degradations.}
\label{tab:robust_degradation}
\scriptsize
\setlength{\tabcolsep}{2.8pt}
\renewcommand{\arraystretch}{1.12}
\begin{tabular*}{\textwidth}{@{\extracolsep{\fill}}lcccccccccccc@{}}
\toprule
\multirow{2}{*}{\textbf{Method}} &
\multirow{2}{*}{\textbf{Sup.}} &
\multirow{2}{*}{\textbf{None}} &
\multicolumn{3}{c}{\textbf{Gaussian Noise ($\sigma$)}} &
\multicolumn{3}{c}{\textbf{Gaussian Blur (Kernel Size)}} &
\multicolumn{3}{c}{\textbf{JPEG Compression}} &
\multirow{2}{*}{\textbf{\shortstack{Avg. Drop\\Rate (\%)}}} \\
\cmidrule(lr){4-6}\cmidrule(lr){7-9}\cmidrule(lr){10-12}
& & & 3 & 7 & 11 & $3\times3$ & $7\times7$ & $11\times11$ & 50 & 70 & 90 & \\
\midrule
% MFI-Net & Full & 0.571 & 0.534 & 0.505 & 0.481 & 0.533 & 0.480 & 0.394 & 0.307 & 0.405 & 0.510 & 19.3 \\
Mesorch & Full & 0.649 & 0.625 & 0.612 & 0.605 & 0.609 & 0.586 & 0.530 & 0.315 & 0.437 & 0.583 & 16.1 \\
PIM & Full & 0.577 & 0.539 & 0.515 & 0.496 & 0.526 & 0.511 & 0.439 & 0.251 & 0.353 & 0.517 & 20.1 \\
PNF-Net & Full & 0.694 & 0.632 & 0.631 & 0.627 & 0.650 & 0.600 & 0.547 & 0.439 & 0.542 & 0.645 & 14.9 \\
Co-Transformers & Full & 0.667 & 0.654 & 0.649 & 0.641 & 0.617 & 0.560 & 0.483 & 0.409 & 0.523 & 0.636 & 13.8 \\
FRD-Net & Full & 0.686 & 0.645 & 0.640 & 0.641 & 0.625 & 0.591 & 0.538 & 0.427 & 0.548 & 0.661 & 13.9 \\
RelayFormer & Full & 0.567 & 0.564 & 0.562 & 0.551 & 0.555 & 0.519 & 0.449 & 0.281 & 0.348 & 0.483 & 15.5 \\
\midrule
\addlinespace[2pt]
\multirow{3}{*}{Ours}
& Scribble & 0.709 & 0.666 & 0.662 & 0.654 & 0.673 & 0.625 & 0.570 & 0.598 & 0.641 & 0.682 & 9.6 \\
& Box & 0.644 & 0.653 & 0.655 & 0.651 & 0.642 & 0.583 & 0.542 & 0.570 & 0.603 & 0.636 & \underline{4.5} \\
& Point & 0.546 & 0.539 & 0.538 & 0.536 & 0.526 & 0.517 & 0.503 & 0.514 & 0.529 & 0.535 & \textbf{3.6} \\
\bottomrule
\end{tabular*}
\vspace{2pt}

\end{table*}

\begin{table}[t]
\centering
\caption{Robustness to online social network processing.}
\label{tab:robust_social}
\setlength{\tabcolsep}{1.4pt}
\renewcommand{\arraystretch}{1.16}
\scriptsize
\begin{tabular*}{\columnwidth}{@{\extracolsep{\fill}}lcccccc@{}}
\toprule
\multirow{2}{*}{\textbf{Method}} & \multirow{2}{*}{\textbf{Sup.}} & \multicolumn{4}{c}{\textbf{Platform}} & \multirow{2}{*}{\textbf{Drop}} \\
\cmidrule(lr){3-6}
& & \textbf{Facebook} & \textbf{WeChat} & \textbf{Weibo} & \textbf{WhatsApp} & \\
\midrule
% MFI-Net & Full & 0.531 & 0.468 & 0.530 & 0.533 & 9.7\% \\
Mesorch & Full & 0.600 & 0.488 & 0.604 & 0.612 & 11.2\% \\
PIM & Full & 0.522 & 0.408 & 0.539 & 0.544 & 12.8\% \\
PNF-Net & Full & 0.627 & 0.483 & 0.620 & 0.633 & 14.9\% \\
RelayFormer & Full & 0.514 & 0.392 & 0.522 & 0.520 & 14.1\% \\
FRD-Net & Full & \underline{0.656} & 0.544 & 0.640 & \underline{0.663} & 8.8\% \\
Co-Transformers & Full & 0.635 & 0.531 & 0.631 & 0.648 & 8.4\% \\\midrule
\addlinespace[2pt]
\multirow{3}{*}{Ours}
& Scribble & \textbf{0.692} & \textbf{0.666} & \textbf{0.695} & \textbf{0.690} & 3.3\% \\
& Box & 0.641 & \underline{0.620} & \underline{0.642} & 0.640 & \textbf{1.3\%} \\
& Point & 0.540 & 0.537 & 0.539 & 0.537 & \underline{1.4\%} \\
\bottomrule
\end{tabular*}
\vspace{2pt}
\end{table}

\subsection{Ablation Studies}
% We analyze the design choices in UniIML across its two distinct stages: OSC-based pseudo-label selection in the first stage and confidence-aware student optimization in the second stage. All ablation variants are evaluated under the same student architecture, training schedule, and evaluation protocol as the full model. 

\subsubsection{Effect of Ordinal Safe Consensus (OSC)}
To evaluate the effectiveness of OSC, we consider three ablation variants under spatial annotation settings: w/o OSC, w/o safety filtering, and w/o ordinal consensus voting. Without OSC, the candidate with the highest SAM-predicted quality score is directly selected from the entire candidate set. Without safety filtering, all candidates participate in the subsequent multi-evidence ordinal ranking and consensus voting. Without ordinal consensus voting, safety filtering is retained, but the final mask is selected solely according to the SAM-predicted quality score among the retained candidates. Image-level supervision requires separate consideration. Positive image-level labels impose only a foreground-presence constraint, without specifying foreground or background pixels or spatial boundaries. Since empty masks are discarded during candidate generation, all remaining candidates satisfy this constraint, and safety filtering therefore leaves the candidate set unchanged. Moreover, removing ordinal consensus voting while retaining safety filtering reduces candidate selection to SAM Top-1 selection, which is identical to the w/o OSC variant under image-level supervision. Therefore, we do not report separate w/o safety filtering and w/o ordinal consensus voting variants for image-level supervision in Table~\ref{tab:ablation_image}.

As shown in Tables~\ref{tab:ablation_spatial} and~\ref{tab:ablation_image}, full OSC achieves higher average localization performance than direct SAM Top-1 selection across all four weak-supervision settings, with improvements in both ID and OOD averages under the three spatial annotation settings. These results suggest that relying solely on SAM-predicted quality scores is insufficient for reliably identifying manipulated regions. Incorporating trusted weak-annotation constraints and multiple sources of evidence into candidate selection helps identify more suitable masks, thereby providing more effective pseudo-label supervision for student training. Under the three spatial annotation settings, removing either safety filtering or ordinal consensus voting reduces average performance, although the magnitude of the degradation varies across supervision settings and evaluation benchmarks. Notably, retaining either component alone generally outperforms direct SAM Top-1 selection in terms of average performance. Under point supervision, the two components contribute comparably to OOD performance. Although individual datasets occasionally favor an ablated variant, full OSC achieves the highest average performance across all four weak-supervision settings, indicating that safety filtering and multi-evidence ordinal consensus voting play complementary roles in candidate mask selection.

Fig.~\ref{fig:osc_case_study} further illustrates the qualitative differences between OSC and direct SAM Top-1 selection. In the second scribble-supervised example, SAM Top-1 produces a large foreground response absent from the ground truth, whereas OSC suppresses this response and retains manipulated regions that better match the reference mask. Under bounding-box supervision, OSC recovers a more complete manipulated region and reduces false foreground responses in authentic regions. Under point supervision, SAM Top-1 tends to select large object regions, while OSC identifies smaller regions that more closely match the ground truth. These cases demonstrate that object masks favored by SAM do not necessarily correspond to manipulated regions. By combining trusted weak-annotation constraints with multi-evidence consensus selection, OSC can identify candidate masks that better align with the requirements of IML.

\subsubsection{Effect of Confidence-Aware Student Optimization}
To evaluate the contributions of confidence-aware student optimization, we consider three ablation variants: w/o confidence weighting, w/o direct supervision, and w/o weighted Dice loss. Without confidence weighting, the teacher confidence weights are replaced with unit weights in regions not covered by direct supervision, affecting both pseudo-label BCE and weighted Dice losses. Without direct supervision, we remove the direct-supervision BCE term and the priority of direct targets, instead using the fixed teacher pseudo-labels and their confidence weights throughout the image. However, the teacher pseudo-labels still retain the annotation information incorporated during their construction. Without the weighted Dice loss, we remove the corresponding loss term from both the main and auxiliary prediction outputs while retaining the BCE objectives.

As shown in Tables~\ref{tab:ablation_spatial}
and~\ref{tab:ablation_image}, the full model achieves the highest average performance across all four weak-supervision settings, although individual datasets exhibit exceptions. Removing confidence weighting reduces average performance under all supervision settings, suggesting that accounting for the varying reliability of teacher-generated pseudo-labels benefits student learning. Removing the weighted Dice loss also leads to performance degradation, particularly on the ID benchmarks under bounding-box supervision, indicating that region-overlap constraints effectively complement pixel-level BCE supervision. In contrast, the effect of direct supervision varies across annotation types. Its removal causes substantial performance degradation under point and image-level supervision, whereas the decreases under scribble and bounding-box supervision are relatively smaller. These results suggest that explicitly utilizing annotation-derived supervision is particularly important when spatial supervision is sparse or absent. Overall, the three optimization components complement each other in improving student learning.

Fig.~\ref{fig:student_ablation_visual} provides a qualitative
comparison under scribble supervision.
Compared with the full model in column (c), removing direct
supervision in column (d) produces more diffuse responses
around the target in the second example and less distinct
boundaries in the fourth.
Without confidence weighting, column (e) shows stronger
background responses and less precise target shapes,
particularly in the first two examples.
Removing the weighted Dice loss in column (f) yields
less complete foreground responses in the third example
and poorer boundary agreement in the fourth.
The full model generally produces more compact predictions
that better match the GT, although errors remain,
including an internal false-negative region in the third example.
These observations complement the quantitative results
and support the combined use of confidence weighting,
direct supervision, and weighted soft Dice supervision.

% \begin{figure*}[!t]
% \centering
% \subfloat[ID Datasets.\label{fig:threshold_stability_ind}]{%
% \includegraphics[width=0.48\textwidth]{threshold_stability_ind_v2.pdf}}\hfill
% \subfloat[OOD Datasets.\label{fig:threshold_stability_ood}]{%
% \includegraphics[width=0.48\textwidth]{threshold_stability_ood_v3.pdf}}
% \caption{Threshold stability of $F_1$ scores on (a) ID and (b) OOD Datasets. The binarization threshold ranges from 0.2 to 0.8. Each curve reports results on one dataset, while the red curve shows the corresponding average.}
% \label{fig:threshold_stability}
% \end{figure*}

\subsection{Robustness Evaluation} 
We compare the robustness of UniIML with several state-of-the-art IML methods. In practical IML, test images often undergo additional post-processing before being analyzed. To evaluate this setting, we test UniIML under two types of distribution changes: synthetic image degradations and online social platform processing. The synthetic setting applies Gaussian blur, Gaussian noise, and JPEG compression to the test images, while the online setting evaluates images processed by Facebook, WeChat, Weibo, and WhatsApp.

As shown in Table~\ref{tab:robust_degradation}, UniIML maintains the highest average $F_1$ score across all synthetic degradation settings. The results indicate that the learned localization cues are relatively stable under common low-level perturbations. Gaussian noise introduces stochastic pixel-level disturbances, but the main manipulated structures are still preserved in the image. JPEG compression changes local statistics through quantization and blocking artifacts, yet manipulation-related inconsistencies can remain distinguishable at the region level. Gaussian blur is more challenging because it suppresses boundary details and high-frequency traces, which are important for precise localization. The performance trend is therefore consistent with the visual nature of these perturbations: degradations that preserve structural inconsistency are easier to tolerate, while degradations that remove fine boundary evidence lead to a narrower localization margin.

Table~\ref{tab:robust_social} further evaluates real post-processing effects introduced by online social platforms. This setting is more practical than isolated synthetic perturbations because platform pipelines may jointly perform resizing, recompression, and format conversion. UniIML remains stable across all four platforms and achieves the best average performance among the compared methods. This suggests that the model does not rely solely on fragile pixel artifacts. Instead, the weakly supervised training encourages the student model to combine local manipulation evidence with spatially coherent region-level supervision, which helps preserve localization ability after real-world dissemination.

Overall, these robustness results provide complementary evidence to the standard benchmark evaluation. UniIML generalizes well under common degradation and platform-processing conditions, showing its potential for practical forensic scenarios. The remaining performance drop under strong blur also reveals a natural limitation: when post-processing removes boundary-level manipulation traces, accurate localization becomes harder even if the manipulated region can still be partially detected.

\section{Conclusion}
This work proposes UniIML, a unified framework for image manipulation localization under multi-type weak supervision. Unlike existing methods that are typically tailored to a specific form of weak annotation, UniIML employs a unified weak-constraint representation to accommodate image-level labels, point annotations, bounding boxes, scribbles, and their mixtures. Within a unified teacher–student framework, heterogeneous weak annotations are further transformed into reliable pixel-level supervision signals. Specifically, we propose Ordinal Safe Consensus (OSC), which combines safety filtering of candidate masks with ordinal consensus voting to reliably select candidate masks while avoiding manually tuned fusion weights across different evidence sources. Based on the selected candidate, UniIML further constructs soft pseudo-labels with pixel-level reliability estimates, while explicitly preserving reliable anchor supervision derived from human-provided weak annotations during student training. This design enables the model to exploit supervision beyond the annotated regions while effectively reducing the adverse influence of unreliable pseudo-labels on the learning process. Extensive experiments demonstrate that UniIML can effectively accommodate different types of weak supervision within a unified localization framework and further support mixed-supervision settings in which multiple forms of weak annotations jointly participate in model training. The experimental results show that UniIML achieves strong IML performance across diverse weak-supervision settings, consistently outperforming the compared weakly supervised baselines. Under several experimental settings, it also matches or surpasses a number of fully supervised methods, while exhibiting favorable cross-dataset generalization performance.

\FloatBarrier

\ifCLASSOPTIONcaptionsoff
  \newpage
\fi
\bibliographystyle{IEEEtran}

\bibliography{reference}

@article{dong2022mvss,
  title={{MVSS-Net}: Multi-view multi-scale supervised networks for image manipulation detection},
  author={Dong, Chengbo and Chen, Xinru and Hu, Ruohan and Cao, Juan and Li, Xirong},
  journal={IEEE Trans. Pattern Anal. Mach. Intell.},
  volume={45},
  number={3},
  pages={3539--3553},
  year={2022},
  publisher={IEEE}
}

@article{liu2024attentive,
  title={Attentive and contrastive image manipulation localization with boundary guidance},
  author={Liu, Wenxi and Zhang, Hao and Lin, Xinyang and Zhang, Qing and Li, Qi and Liu, Xiaoxiang and Cao, Ying},
  journal={IEEE Trans. Inf. Forensics Security},
  volume={19},
  pages={6764--6778},
  year={2024},
  publisher={IEEE}
}

@inproceedings{li2024unionformer,
  title={{UnionFormer}: Unified-learning transformer with multi-view representation for image manipulation detection and localization},
  author={Li, Shuaibo and Ma, Wei and Guo, Jianwei and Xu, Shibiao and Li, Benchong and Zhang, Xiaopeng},
  booktitle={Proc. IEEE/CVF Conf. Comput. Vis. Pattern Recognit. (CVPR)},
  pages={12523--12533},
  year={2024}
}

@article{li2025bridging,
  title={Bridging Semantic Logic Gaps: A Cognition Inspired Multimodal Boundary Preserving Network for Image Manipulation Localization},
  author={Li, Songlin and Guo, Zhiqing and Li, Yuanman and Li, Zeyu and Diao, Yunfeng and Yang, Gaobo and Wang, Liejun},
  journal={arXiv preprint arXiv:2508.07216},
  year={2025}
}

@article{niu2026dr,
  title={{DR-Net}: Depth-assisted refinement network for image manipulation localization},
  author={Niu, Yiqian and Zhu, Xiaodong and Shi, Zenan and Lyu, Yingda},
  journal={Neurocomputing},
  note={{Art. no.}~133776},
  year={2026},
  publisher={Elsevier}
}

@inproceedings{zhai2023towards,
  title={Towards Generic Image Manipulation Detection with Weakly-Supervised Self-Consistency Learning},
  author={Zhai, Yuanhao and Luan, Tianyu and Doermann, David and Yuan, Junsong},
  booktitle={Proc. IEEE/CVF Int. Conf. Comput. Vis. (ICCV)},
  pages={22390--22400},
  year={2023}
}

@article{ZHOU2024123501,
title = {Exploring weakly-supervised image manipulation localization with tampering Edge-based class activation map},
journal = {Expert Syst. Appl.},
volume = {249},
note={{Art. no.}~123501},
year = {2024},
issn = {0957-4174},
doi = {10.1016/j.eswa.2024.123501},
author = {Yang Zhou and Hongxia Wang and Qiang Zeng and Rui Zhang and Sijiang Meng}
}

@INPROCEEDINGS{10889843,
  author={Zhu, Zhangchen and Li, Jiafeng and Wen, Ying},
  booktitle={Proc. IEEE Int. Conf. Acoust., Speech Signal Process. (ICASSP)},
  title={Self-Optimization Training for Weakly Supervised Image Manipulation Localization},
  year={2025},
  pages={1--5},
  doi={10.1109/ICASSP49660.2025.10889843}}

@article{bai2025weakly,
  title={Weakly-supervised cross-contrastive learning network for image manipulation detection and localization},
  author={Bai, Ruyi},
  journal={Knowl.-Based Syst.},
  volume={310},
  note={{Art. no.}~113033},
  year={2025},
  publisher={Elsevier}
}

@article{wang2025context,
  title={Context-Aware Weakly Supervised Image Manipulation Localization with {SAM} Refinement},
  author={Wang, Xinghao and Miao, Changtao and Sheng, Dianmo and Gong, Tao and Chu, Qi and Liu, Bin and Yu, Nenghai},
  journal={arXiv preprint arXiv:2503.20294},
  year={2025}
}

@inproceedings{sheng2026weakly,
  title={Weakly-Supervised Image Forgery Localization via Vision-Language Collaborative Reasoning Framework},
  author={Sheng, Ziqi and Wu, Junyan and Lu, Wei and Zhou, Jiantao},
  booktitle={Proc. AAAI Conf. Artif. Intell.},
  volume={40},
  number={2},
  pages={988--996},
  year={2026},
  doi={10.1609/aaai.v40i2.37068}
}

@inproceedings{guo2026passive,
  title={From Passive Perception to Active Memory: A Weakly Supervised Image Manipulation Localization Framework Driven by Coarse-Grained Annotations},
  author={Guo, Zhiqing and Xi, Dongdong and Li, Songlin and Yang, Gaobo},
  booktitle={Proc. AAAI Conf. Artif. Intell.},
  volume={40},
  number={1},
  pages={336--344},
  year={2026},
  doi={10.1609/aaai.v40i1.36995}
}

@inproceedings{li2026beyond,
  title={Beyond fully supervised pixel annotations: Scribble-driven weakly-supervised framework for image manipulation localization},
  author={Li, Songlin and Yu, Guofeng and Guo, Zhiqing and Diao, Yunfeng and Ma, Dan and Yang, Gaobo},
  booktitle={Proc. AAAI Conf. Artif. Intell.},
  volume={40},
  number={1},
  pages={677--685},
  year={2026},
  doi={10.1609/aaai.v40i1.37033}
}

@inproceedings{kirillov2023segment,
  title={Segment anything},
  author={Kirillov, Alexander and Mintun, Eric and Ravi, Nikhila and Mao, Hanzi and Rolland, Chloe and Gustafson, Laura and Xiao, Tete and Whitehead, Spencer and Berg, Alexander C and Lo, Wan-Yen and others},
  booktitle={Proc. IEEE/CVF Int. Conf. Comput. Vis. (ICCV)},
  pages={4015--4026},
  year={2023}
}

@inproceedings{guillaro2023trufor,
  title={{TruFor}: Leveraging all-round clues for trustworthy image forgery detection and localization},
  author={Guillaro, Fabrizio and Cozzolino, Davide and Sud, Avneesh and Dufour, Nicholas and Verdoliva, Luisa},
  booktitle={Proc. IEEE/CVF Conf. Comput. Vis. Pattern Recognit. (CVPR)},
  pages={20606--20615},
  year={2023}
}

@ARTICLE{mfi2024,
  author={Ren, Ruyong and Hao, Qixian and Niu, Shaozhang and Xiong, Keyang and Zhang, Jiwei and Wang, Maosen},
  journal={IEEE Trans. Circuits Syst. Video Technol.},
  title={{MFI-Net}: Multi-Feature Fusion Identification Networks for Artificial Intelligence Manipulation},
  year={2024},
  volume={34},
  number={2},
  pages={1266--1280},
  doi={10.1109/TCSVT.2023.3289171}}

@inproceedings{zhu2025mesoscopic,
  title={Mesoscopic insights: orchestrating multi-scale \& hybrid architecture for image manipulation localization},
  author={Zhu, Xuekang and Ma, Xiaochen and Su, Lei and Jiang, Zhuohang and Du, Bo and Wang, Xiwen and Lei, Zeyu and Feng, Wentao and Pun, Chi-Man and Zhou, Ji-Zhe},
  booktitle={Proc. AAAI Conf. Artif. Intell.},
  volume={39},
  number={10},
  pages={11022--11030},
  year={2025}
}

@article{li2026prototype,
  title={Prototype Memory-based Neighboring Feature Fusion Network for Image Manipulation Localization},
  author={Li, Songlin and Guo, Zhiqing and Miao, Changtao and Yang, Wenzhong and Wang, Liejun and Yang, Gaobo and Liao, Xin},
  journal={IEEE Trans. Circuits Syst. Video Technol.},
  year={2026},
  publisher={IEEE}
}

@inproceedings{dong2013casia,
  title={{CASIA} Image Tampering Detection Evaluation Database},
  author={Dong, Jing and Wang, Wei and Tan, Tieniu},
  booktitle={Proc. IEEE China Summit Int. Conf. Signal Inf. Process. (ChinaSIP)},
  pages={422--426},
  year={2013},
  doi={10.1109/ChinaSIP.2013.6625374}
}

@inproceedings{wen2016coverage,
  title={{COVERAGE}---A Novel Database for Copy-Move Forgery Detection},
  author={Wen, Bihan and Zhu, Ye and Subramanian, Ramanathan and Ng, Tian-Tsong and Shen, Xuanjing and Winkler, Stefan},
  booktitle={Proc. IEEE Int. Conf. Image Process. (ICIP)},
  pages={161--165},
  year={2016},
  doi={10.1109/ICIP.2016.7532339}
}

@inproceedings{ma2024imdlbenco,
  title={{IMDL-BenCo}: A Comprehensive Benchmark and Codebase for Image Manipulation Detection \& Localization},
  author={Ma, Xiaochen and Zhu, Xuekang and Su, Lei and Du, Bo and Jiang, Zhuohang and Tong, Bingkui and Lei, Zeyu and Yang, Xinyu and Pun, Chi-Man and Lv, Jiancheng and Zhou, Jizhe},
  booktitle={Adv. Neural Inf. Process. Syst.},
  volume={37},
  year={2024}
}

@inproceedings{su2025sparsevit,
  title={Can we get rid of handcrafted feature extractors? {SparseViT}: Nonsemantics-centered, parameter-efficient image manipulation localization through sparse-coding transformer},
  author={Su, Lei and Ma, Xiaochen and Zhu, Xuekang and Niu, Chaoqun and Lei, Zeyu and Zhou, Ji-Zhe},
  booktitle={Proc. AAAI Conf. Artif. Intell.},
  volume={39},
  number={7},
  pages={7024--7032},
  year={2025},
  doi={10.1609/aaai.v39i7.32754}
}

@article{bai2025pimnet,
  title={{PIM-Net}: Progressive inconsistency mining network for image manipulation localization},
  author={Bai, Ningning and Wang, Xiaofeng and Han, Ruidong and Hou, Jianpeng and Wang, Yihang and Pang, Shanmin},
  journal={Pattern Recognit.},
  volume={159},
  year={2025},
  note={{Art. no.}~111136},
  doi={10.1016/j.patcog.2024.111136}
}

@article{kong2025pixel,
  title={Pixel-inconsistency modeling for image manipulation localization},
  author={Kong, Chenqi and Luo, Anwei and Wang, Shiqi and Li, Haoliang and Rocha, Anderson and Kot, Alex C.},
  journal={IEEE Trans. Pattern Anal. Mach. Intell.},
  year={2025},
  doi={10.1109/TPAMI.2025.3541028},
  note={doi:~10.1109/TPAMI.2025.3541028}
}

@inproceedings{zhang2026collaborative,
  title={Collaborative transformers with multi-level forensic attention for image manipulation localization},
  author={Zhang, Jiwei and Feng, Wenbo and Wang, Siwei and Kou, Feifei and Yu, Haoyang and Niu, Shaozhang},
  booktitle={Proc. AAAI Conf. Artif. Intell.},
  volume={40},
  number={15},
  pages={12556--12563},
  year={2026},
  doi={10.1609/aaai.v40i15.38250}
}

@inproceedings{chen2026amplifying,
  title={Amplifying discrepancies: Exploiting macro and micro inconsistencies for image manipulation localization},
  author={Chen, Shenghao and Zhao, Yibo and Wang, Tianyi and Ma, Chunjie and Guan, Weili and Li, Ming and Gao, Zan},
  booktitle={Proc. AAAI Conf. Artif. Intell.},
  volume={40},
  number={42},
  pages={35357--35365},
  year={2026},
  doi={10.1609/aaai.v40i42.40844}
}

@inproceedings{huang2026relayformer,
  title={{RelayFormer}: A unified local-global attention framework for scalable image and video manipulation localization},
  author={Huang, Wen and Yang, Jiarui and Dai, Tao and Li, Jiawei and Zhan, Shaoxiong and Wang, Bin and Xia, Shu-Tao},
  booktitle={Proc. Int. Conf. Learn. Represent. (ICLR)},
  year={2026}
}

@IEEEtranBSTCTL{uniml:bstcontrol,
  CTLuse_forced_etal = {yes},
  CTLmax_names_forced_etal = {6},
  CTLnames_show_etal = {1},
  CTLdash_repeated_names = {no}
}

@inproceedings{guan2019mfc,
  title={MFC datasets: Large-scale benchmark datasets for media forensic challenge evaluation},
  author={Guan, Haiying and Kozak, Mark and Robertson, Eric and Lee, Yooyoung and Yates, Amy N and Delgado, Andrew and Zhou, Daniel and Kheyrkhah, Timothee and Smith, Jeff and Fiscus, Jonathan},
  booktitle={2019 IEEE Winter Applications of Computer Vision Workshops (WACVW)},
  pages={63--72},
  year={2019},
  organization={IEEE}
}

@article{hsu2006columbia,
  title={Columbia uncompressed image splicing detection evaluation dataset},
  author={Hsu, J and Chang, SF},
  journal={Columbia DVMM Research Lab},
  volume={6},
  year={2006}
}

@inproceedings{korus2016evaluation,
  title={Evaluation of random field models in multi-modal unsupervised tampering localization},
  author={Korus, Pawe{\l} and Huang, Jiwu},
  booktitle={2016 IEEE international workshop on information forensics and security (WIFS)},
  pages={1--6},
  year={2016},
  organization={IEEE}
}

@article{de2013exposing,
  title={Exposing digital image forgeries by illumination color classification},
  author={De Carvalho, Tiago Jos{\'e} and Riess, Christian and Angelopoulou, Elli and Pedrini, Helio and de Rezende Rocha, Anderson},
  journal={IEEE Transactions on Information Forensics and Security},
  volume={8},
  number={7},
  pages={1182--1194},
  year={2013},
  publisher={IEEE}
}

@INPROCEEDINGS{Novozamsky_2020_WACV,
author = {Novozamsky, Adam and Mahdian, Babak and Saic, Stanislav},
title = {IMD2020: A Large-Scale Annotated Dataset Tailored for Detecting Manipulated Images},
booktitle = {2020 IEEE Winter Applications of Computer Vision Workshops (WACVW)},
year = {2020},
month = {March},
pages = {71-80}
}

@article{nichol2021glide,
  title={Glide: Towards photorealistic image generation and editing with text-guided diffusion models},
  author={Nichol, Alex and Dhariwal, Prafulla and Ramesh, Aditya and Shyam, Pranav and Mishkin, Pamela and McGrew, Bob and Sutskever, Ilya and Chen, Mark},
  journal={arXiv preprint arXiv:2112.10741},
  year={2021}
}

@inproceedings{huh2018fighting,
  title={Fighting fake news: Image splice detection via learned self-consistency},
  author={Huh, Minyoung and Liu, Andrew and Owens, Andrew and Efros, Alexei A},
  booktitle={Proceedings of the European conference on computer vision (ECCV)},
  pages={101--117},
  year={2018}
}

@article{wang2022pvt,
  title={Pvt v2: Improved baselines with pyramid vision transformer},
  author={Wang, Wenhai and Xie, Enze and Li, Xiang and Fan, Deng-Ping and Song, Kaitao and Liang, Ding and Lu, Tong and Luo, Ping and Shao, Ling},
  journal={Computational Visual Media},
  volume={8},
  number={3},
  pages={415--424},
  year={2022},
  publisher={Springer}
}

@inproceedings{loshchilovdecoupled,
  title={Decoupled Weight Decay Regularization},
  author={Loshchilov, Ilya and Hutter, Frank},
  booktitle={International Conference on Learning Representations},
  year={2019}
}

@article{verdoliva2020media,
  title={Media forensics and deepfakes: an overview},
  author={Verdoliva, Luisa},
  journal={IEEE journal of selected topics in signal processing},
  volume={14},
  number={5},
  pages={910--932},
  year={2020},
  publisher={IEEE}
}

@article{mirsky2021creation,
  title={The creation and detection of deepfakes: A survey},
  author={Mirsky, Yisroel and Lee, Wenke},
  journal={ACM computing surveys (CSUR)},
  volume={54},
  number={1},
  pages={1--41},
  year={2021},
  publisher={ACM New York, NY, USA}
}

@article{kadha2025unravelling,
  title={Unravelling digital forgeries: A systematic survey on image manipulation detection and localization},
  author={Kadha, VijayaKumar and Bakshi, Sambit and Das, Santos Kumar},
  journal={ACM Computing Surveys},
  volume={57},
  number={12},
  pages={1--36},
  year={2025},
  publisher={ACM New York, NY}
}

@article{zhang2026noise,
  title={Noise-aware cross attention for image manipulation localization},
  author={Zhang, Hongshi and Su, Tonghua and Liu, Zhou and Yang, Fuxiang and Di, Donglin and Song, Yang and Fan, Lei},
  journal={Pattern Recognition},
  pages={113164},
  year={2026},
  publisher={Elsevier}
}

@inproceedings{yu2026earg,
  title={Earg-net: edge-aware reconstruction-guided network for image manipulation detection and localization},
  author={Yu, Yanpu and Shi, Zhaoxin and Zhao, Hanqing and Wei, Tianyi and Zhou, Wenbo and Yu, Nenghai},
  booktitle={Proceedings of the AAAI Conference on Artificial Intelligence},
  volume={40},
  number={14},
  pages={12178--12186},
  year={2026}
}

@inproceedings{lin2016scribblesup,
  title={Scribblesup: Scribble-supervised convolutional networks for semantic segmentation},
  author={Lin, Di and Dai, Jifeng and Jia, Jiaya and He, Kaiming and Sun, Jian},
  booktitle={Proceedings of the IEEE conference on computer vision and pattern recognition},
  pages={3159--3167},
  year={2016}
}

@article{bai2025qwen3,
  title={Qwen3-vl technical report},
  author={Bai, Shuai and Cai, Yuxuan and Chen, Ruizhe and Chen, Keqin and Chen, Xionghui and Cheng, Zesen and Deng, Lianghao and Ding, Wei and Gao, Chang and Ge, Chunjiang and others},
  journal={arXiv preprint arXiv:2511.21631},
  year={2025}
}

@article{yu20262d,
  title={When 2D Cues Fail: Improving Image Manipulation Localization with Reliable 3D Geometry},
  author={Yu, Guofeng and Guo, Zhiqing and Ma, Dan and Yang, Gaobo},
  journal={arXiv preprint arXiv:2607.18040},
  year={2026}
}

@inproceedings{wu2019mantra,
  title={Mantra-net: Manipulation tracing network for detection and localization of image forgeries with anomalous features},
  author={Wu, Yue and AbdAlmageed, Wael and Natarajan, Premkumar},
  booktitle={2019 IEEE/CVF Conference on Computer Vision and Pattern Recognition (CVPR)},
  pages={9535--9544},
  year={2019},
  organization={IEEE}
}

@inproceedings{kwon2021cat,
  title={CAT-Net: Compression artifact tracing network for detection and localization of image splicing},
  author={Kwon, Myung-Joon and Yu, In-Jae and Nam, Seung-Hun and Lee, Heung-Kyu},
  booktitle={2021 IEEE winter conference on applications of computer vision (WACV)},
  pages={375--384},
  year={2021},
  organization={IEEE}
}

@article{liu2022pscc,
  title={PSCC-Net: Progressive spatio-channel correlation network for image manipulation detection and localization},
  author={Liu, Xiaohong and Liu, Yaojie and Chen, Jun and Liu, Xiaoming},
  journal={IEEE Transactions on Circuits and Systems for Video Technology},
  volume={32},
  number={11},
  pages={7505--7517},
  year={2022},
  publisher={IEEE}
}

@inproceedings{wang2022objectformer,
  title={Objectformer for image manipulation detection and localization},
  author={Wang, Junke and Wu, Zuxuan and Chen, Jingjing and Han, Xintong and Shrivastava, Abhinav and Lim, Ser-Nam and Jiang, Yu-Gang},
  booktitle={2022 IEEE/CVF Conference on Computer Vision and Pattern Recognition (CVPR)},
  pages={2354--2363},
  year={2022},
  organization={IEEE}
}

@inproceedings{hu2020span,
  title={SPAN: Spatial pyramid attention network for image manipulation localization},
  author={Hu, Xuefeng and Zhang, Zhihan and Jiang, Zhenye and Chaudhuri, Syomantak and Yang, Zhenheng and Nevatia, Ram},
  booktitle={European conference on computer vision},
  pages={312--328},
  year={2020},
  organization={Springer}
}

@inproceedings{sudre2017generalised,
  title={Generalised dice overlap as a deep learning loss function for highly unbalanced segmentations},
  author={Sudre, Carole H and Li, Wenqi and Vercauteren, Tom and Ourselin, Sebastien and Jorge Cardoso, M},
  booktitle={International Workshop on Deep Learning in Medical Image Analysis},
  pages={240--248},
  year={2017},
  organization={Springer}
}

\begin{IEEEbiography}[{\includegraphics[width=1.0in,height=1.2in,clip,keepaspectratio]{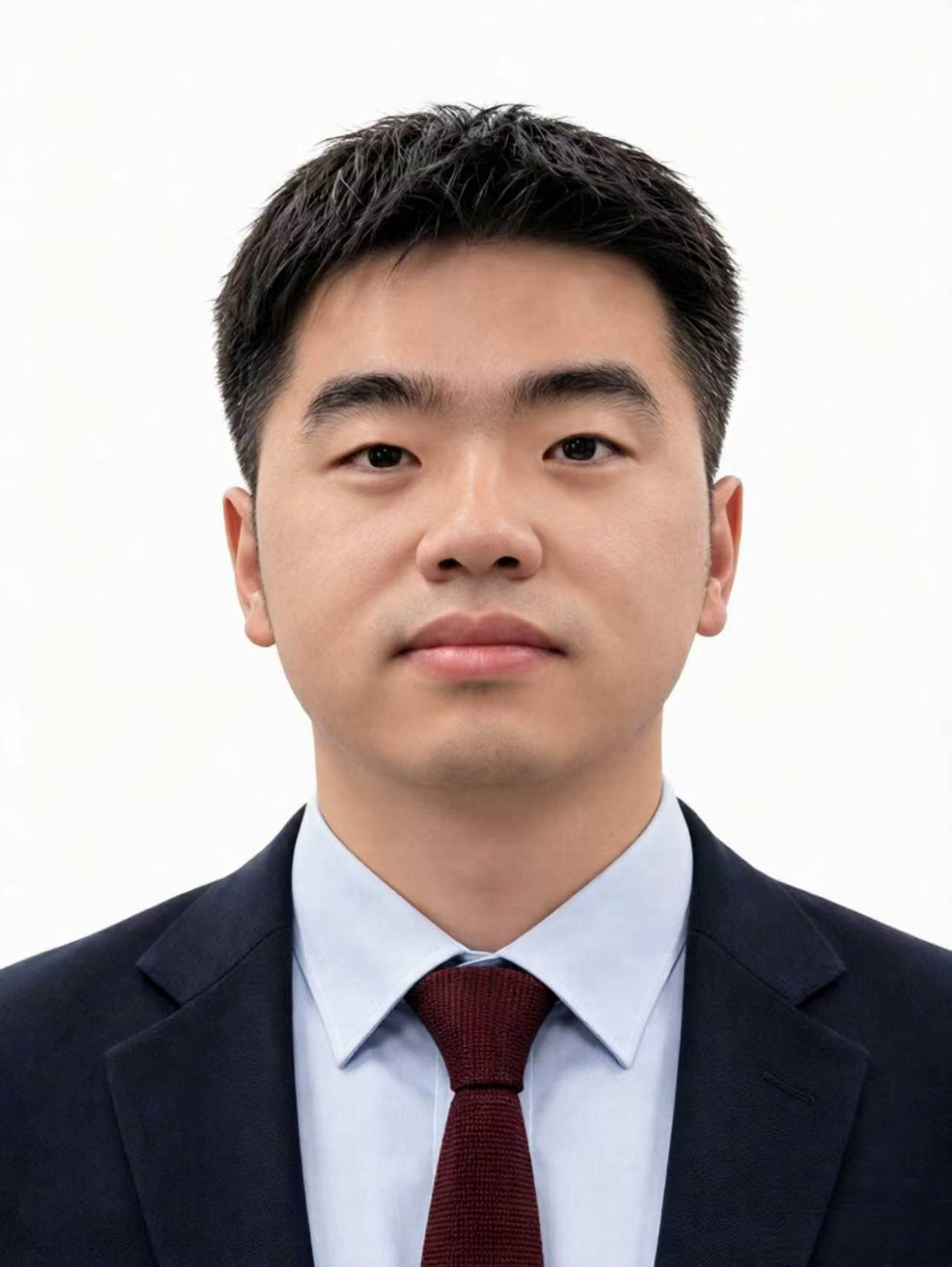}}]{Songlin~Li} is currently pursuing a Ph.D. degree in Computer Science and Technology at the School of Computer Science and Technology, Xinjiang University, China. His current research interests include co-camouflage object detection, image manipulation detection and deep learning. He has published more than 10 papers in international journals and conferences such as Proceedings of the AAAI Conference on Artificial Intelligence (AAAI), IEEE Transactions on Circuits and Systems for Video Technology (TCSVT), the IEEE International Conference on Multimedia \& Expo (ICME), and the International Conference on Multimedia Retrieval (ICMR), etc.
\end{IEEEbiography}

\begin{IEEEbiography}[{\includegraphics[width=1.0in,height=1.2in,clip,keepaspectratio]{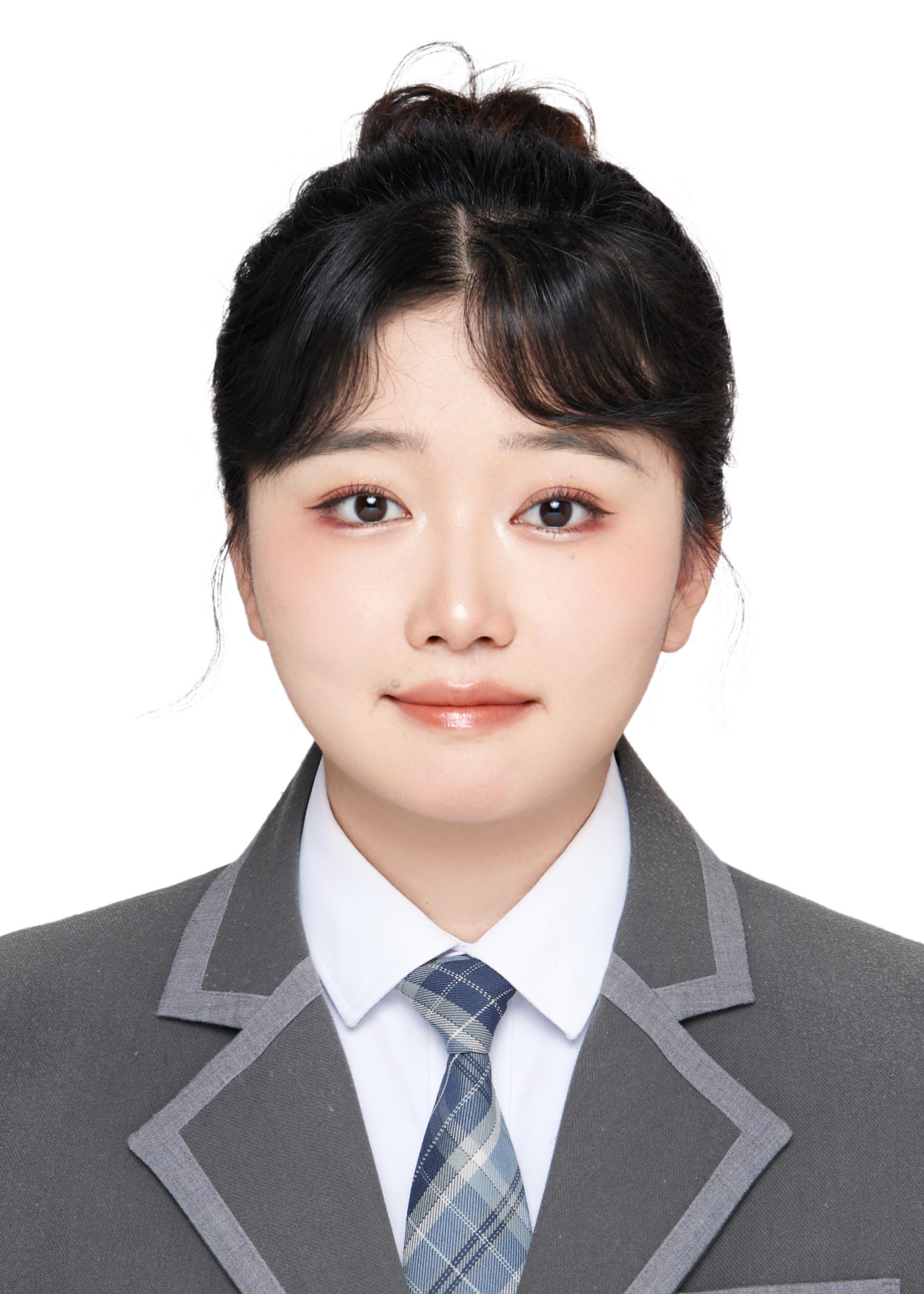}}]{Hui~Ma} is currently pursuing the M.S. degree with the School of Computer Science and Technology, Xinjiang University, Urumqi, China. Her current research interests include digital media forensics, image manipulation localization, and deep learning, with a particular focus on robust and accurate localization of manipulated regions in digital images.
\end{IEEEbiography}

\begin{IEEEbiography}[{\includegraphics[width=1.0in,height=1.2in,clip,keepaspectratio]{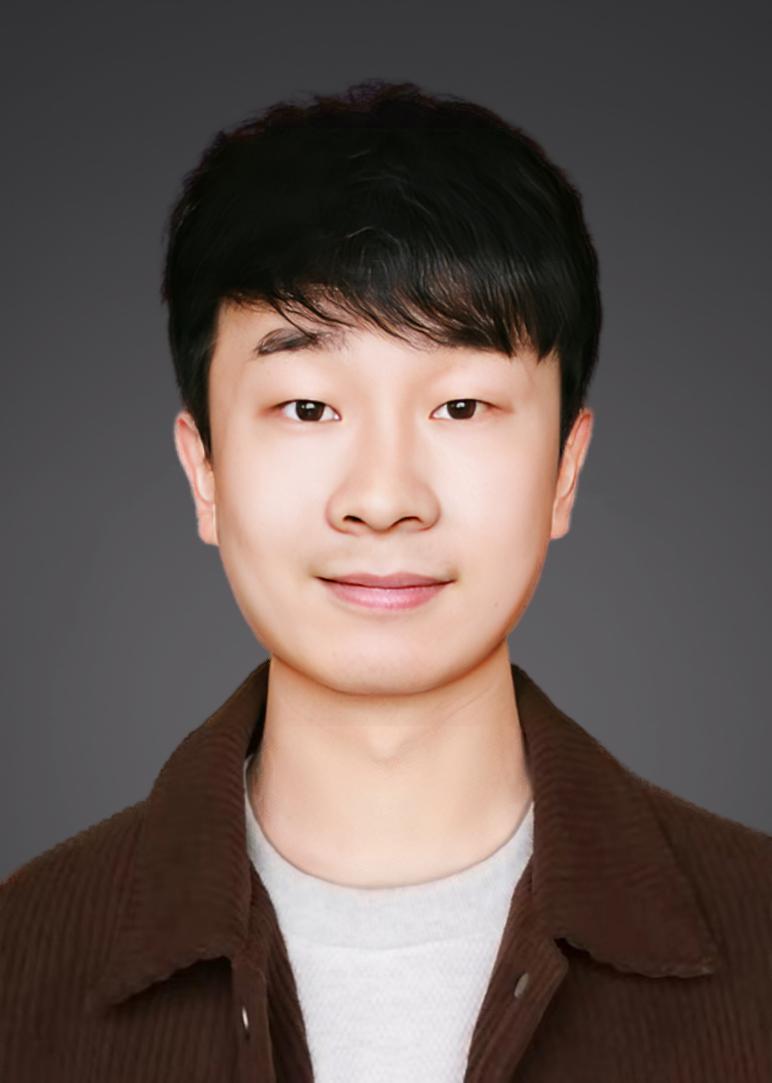}}]{Zhiqing~Guo (Member, IEEE)} received the Ph.D. degree in Computer Science and Technology from Hunan University in 2023. He is currently an Associate Professor at the School of Computer Science and Technology, Xinjiang University, Urumqi, China. At present, he is the PI of several projects, such as the Natural Science Foundation of China (NSFC), with a total fund of more than 2 million yuan. His research interests include multimedia forensics, image/video content security, visual scene understanding. He has published more than 50 papers in international journals or conferences such as IEEE Transactions on Information Forensics and Security (TIFS), IEEE Transactions on Multimedia (TMM), IEEE Transactions on Circuits and Systems for Video Technology (TCSVT), ACM Transactions on Multimedia Computing Communications and Applications (TOMM), International Conference on Machine Learning (ICML), IEEE/CVF Conference on Computer Vision and Pattern Recognition (CVPR), International Conference on Computer Vision (ICCV), Proceedings of the AAAI Conference on Artificial Intelligence (AAAI), etc.
\end{IEEEbiography}

\begin{IEEEbiography}[{\includegraphics[width=1.0in,height=1.2in,clip,keepaspectratio]{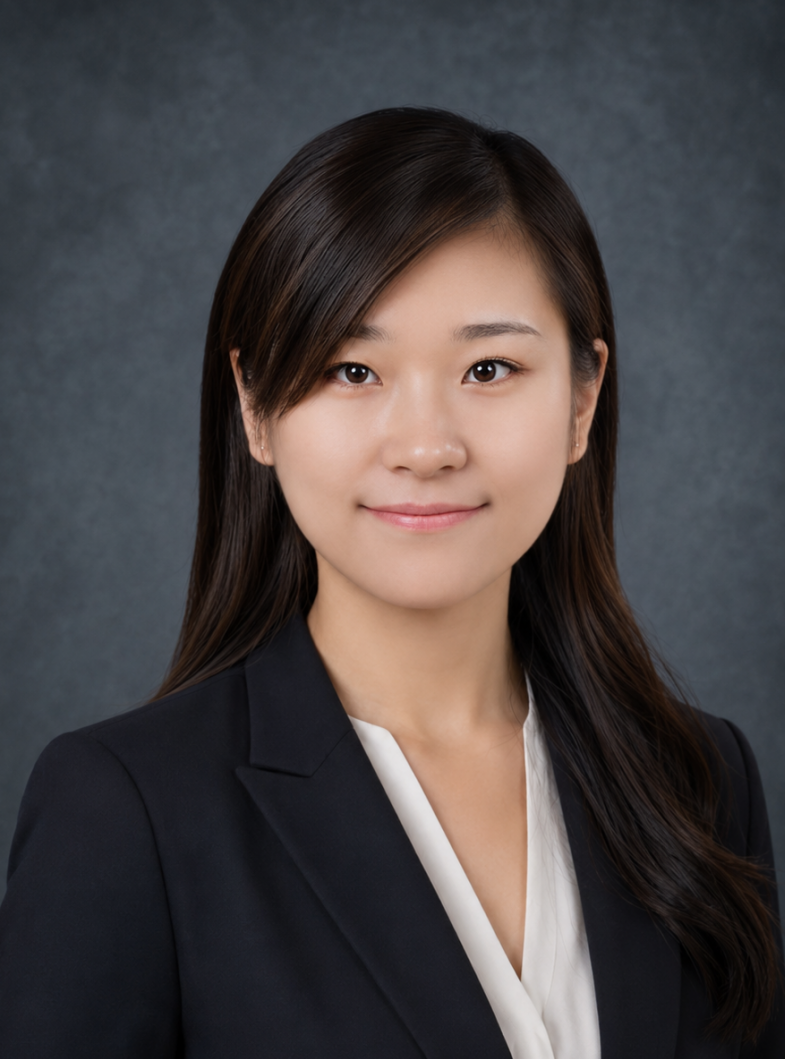}}]{Dan~Ma (Member, IEEE)} received the Ph.D. degree in Information and Communication Systems from Tianjin University, Tianjin, China, in 2025. She is currently a Master's Supervisor at the School of Computer Science and Technology. Her research focuses on remote sensing intelligent interpretation, trustworthy multimodal analysis, and content security. She has published over 20 papers in high-impact journals and conferences such as IEEE TGRS, IEEE TCSVT, INFFUS, AAAI, and ACM MM, with a Best Paper Nomination at the IJCAI Workshop. She holds multiple authorized invention patents and serves as a reviewer for international journals and conferences including IEEE JSTARS and AAAI. She has also participated in more than ten national and provincial research projects, including major programs of the National Natural Science Foundation of China and the National Key R\&D Program.
\end{IEEEbiography}

\begin{IEEEbiography}[{\includegraphics[width=1.0in,height=1.2in,clip,keepaspectratio]{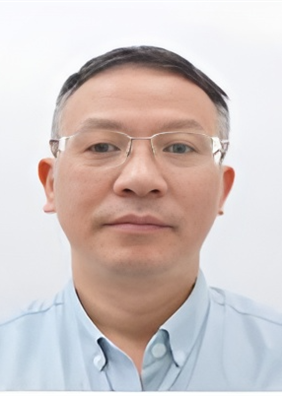}}]{Gaobo~Yang} received the Ph.D. degree in Communication and Information System from Shanghai University in 2004. He is a professor in Hunan University, China. He made an academic visit to University of Surrey, UK, from August 2010 to August 2011. He is the PI of several projects such as Natural Science Foundation of China (NSFC), Special Pro-phase Project on National Basic Research Program of China (973) and program for New Century Excellent Talents (NCET) in university. His research interests include image and video signal processing, digital media forensics. He has published research papers in international journals such as IEEE Transactions on Multimedia, IEEE Transactions on Circuits and Systems for Video Technology, IEEE Transactions on Broadcasting, ACM Transactions on Multimedia Computing, Communications, and Applications, etc.
\end{IEEEbiography}

\begin{IEEEbiography}[{\includegraphics[width=1.0in,height=1.2in,clip,keepaspectratio]{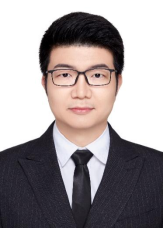}}]{Xin~Liao (Senior Member, IEEE)} received the B.E. and Ph.D. degrees in information security from the Beijing University of Posts and Telecommunications in 2007 and 2012, respectively. He is currently a Professor with the College of Cyber Science and Technology, Hunan University, where he also serves as the Vice Dean. He worked as a Post-Doctoral Fellow with the Institute of Software, Chinese Academy of Sciences, and also a Research Associate with The University of Hong Kong. From 2016 to 2017, he was a Visiting Scholar with the University of Maryland, College Park, USA. His current research interests include multimedia forensics, AI security, and watermarking. He is the Vice Chair of Technical Committee (TC) on Multimedia Security and Forensics of Asia-Pacific Signal and Information Processing Association, a member of TC on Computer Forensics of the Chinese Institute of Electronics, and a member of TC on Digital Forensics and Security of the China Society of Image and Graphics. He is serving as an Associate Editor for the IEEE Signal Processing Magazine.
\end{IEEEbiography}

\end{document}